\documentclass[journal]{IEEEtran}

\usepackage{hhline}
\usepackage{color, colortbl}
\usepackage{multirow}
\usepackage{times}
\usepackage{epsfig}
\usepackage{graphicx}
\usepackage{amsmath}
\usepackage{amssymb}
\usepackage{subcaption}
\usepackage[utf8]{inputenc}
\usepackage{lipsum}
\usepackage{booktabs}
\usepackage{xcolor}
\usepackage{makecell}
\usepackage{url}
\usepackage{authblk}

\usepackage[bookmarks=false]{hyperref}

\definecolor{lightgray}{gray}{0.9}
\definecolor{lightblue}{rgb}{0.93,0.95,1.0}
\definecolor{darkgreen}{rgb}{0.0,0.6,0.0}
\definecolor{blue}{rgb}{1, 0, 0}

\makeatletter

\def\ps@IEEEtitlepagestyle{
\def\@oddfoot{\mycopyrightnotice}
\def\@evenfoot{}
}
\def\mycopyrightnotice{
{\hfill \footnotesize \copyright 2021 IEEE. Personal use of this material is permitted. Permission from IEEE must be obtained for all other uses, in any current or future media, including reprinting/republishing this material for advertising or promotional purposes, creating new collective works, for resale or redistribution to servers or lists, or reuse of any copyrighted component of this work in other works.\hfill
\hfill \footnotesize DOI: \href{https://doi.org/10.1109/TIFS.2021.3135750}{10.1109/TIFS.2021.3135750}\hfill}
}

\makeatother

\begin{document}

\title{Gendered Differences in Face Recognition Accuracy Explained by Hairstyles, Makeup, and Facial Morphology}

\author[1]{Vítor Albiero,~\IEEEmembership{Graduate Student Member,~IEEE}}
\author[1]{Kai Zhang}
\author[2]{Michael C. King,~\IEEEmembership{Member,~IEEE}}
\author[1]{Kevin W. Bowyer,~\IEEEmembership{Fellow,~IEEE}}
\affil[1]{University of Notre Dame, Notre Dame, Indiana}
\affil[2]{Florida Institute of Technology, Melbourne, Florida}

\markboth{PREPRINT SUBMITTED TO IEEE TRANSACTIONS ON INFORMATION FORENSICS AND SECURITY}%
{Shell \MakeLowercase{\textit{et al.}}: Bare Demo of IEEEtran.cls for IEEE Journals}
%

\maketitle
\thispagestyle{empty}

\begin{abstract}
Media reports have accused face recognition of being ``biased'', ``sexist'' and ``racist''.
There is consensus in the research literature that face recognition accuracy is lower for females, who often have both a higher false match rate and a higher false non-match rate.
However, there is little published research aimed at identifying the cause of lower accuracy for females.
For instance, the 2019 Face Recognition Vendor Test that documents lower female accuracy across a broad range of algorithms and datasets also lists
``Analyze cause and effect’’ under the heading ``What we did not do’’.
We present the first experimental analysis to identify major causes of lower face recognition accuracy for females on datasets where previous research has observed this result.
Controlling for equal amount of visible face in the test images mitigates the apparent higher false non-match rate for females. 
Additional analysis shows that makeup-balanced datasets further improves females to achieve lower false non-match rates.
Finally, a clustering experiment suggests that images of two different females are inherently more similar than of two different males, potentially accounting for a difference in false match rates.
\end{abstract}

\begin{IEEEkeywords}
Face recognition, gender, fairness, bias, hairstyle, makeup, facial morphology.
\end{IEEEkeywords}

{\let\thefootnote\relax\footnotetext{\mycopyrightnotice}}

\IEEEpeerreviewmaketitle

\section{Introduction}

Many recent news articles have criticized face recognition 
as being ``biased'', ``sexist'' or ``racist''~\cite{Lohr2018,Hoggins2019,Doctorow2019,Santow2020}.
Various papers summarized in Related Work report an empirical result
that face recognition is less accurate for females.
Accuracy also varies between racial groups, and age ranges, but this paper focuses on the female / male difference.
Surprisingly, 
given the controversy generated by the issue of unequal accuracy, the underlying cause(s) of lower accuracy for females are heretofore unknown.
The popularity of deep CNN algorithms perhaps naturally gives rise to the common 
speculation that the cause is female under-representation in the training data.
But recent studies show that simply using a gender-balanced training set does not result in balanced accuracy on test data~\cite{Albiero2020_gender, Albiero2020_train}. 

This paper extends our previous work~\cite{albiero2020skin} presenting experimental results that explain the causes of lower face recognition accuracy for females.
In this paper, we expand the results to include a COTS face matcher, a face matcher trained with gender-balanced training set, a new dataset with a different ethnicity (Asian) and type (web-scraped), an analysis of the effects of makeup, and a clustering-based explanation for differences in the impostor distribution.
Our experiments show that (1) gendered hairstyles result in, on average, more of the face being occluded for females, (2) makeup use degrades the female genuine distribution, and (3) the inherent variability between different female faces is lower than for males, leading to a worse impostor distribution.

To promote reproducible and transparent results, we use state-of-the-art deep CNN matchers and datasets that are available to other researchers.
To better ensure that our results are generally applicable, we also experiment with a third commercial-of-the-shelf (COTS) matcher.
In summary, the contributions of this paper are:
\begin{itemize}
    \item Curation of a web-scraped Asian dataset;
    \item Extension of female/male accuracy analysis to an Asian dataet;
    \item Analysis of the effect of skin visibility, due to a combination of gendered hairstyles and face size/shape difference, on female/male accuracy;
    \item Clustering analysis to compare the inherent similarity of face appearance across a set of different females versus different males.
\end{itemize}

\section{Related Work}
\label{sec:related_work}

Drozdowski et al.~\cite{Drozdowski2020} give a broad survey of current work related to demographic bias in biometrics. 
Here we focus on selected prior works dealing specifically with female / male difference in face recognition accuracy.

\subsection{Fairness in Face Recognition Before Deep Learning}

The earliest work we are aware of to report lower accuracy for females is the 2002 Face Recognition Vendor Test (FRVT)~\cite{frvt}. 
Evaluating ten algorithms of that era, 
identification rates of the top systems are 6\% to 9\% lower for females.
However, for the highest-accuracy matcher, accuracy was essentially the same for females and males. 
In a 2009 meta-analysis of results from eight prior papers, Lui et al.~\cite{Lui2009} 
concluded that there was a weak pattern of lower accuracy for females, and noted interactions with factors such as age, expression, lighting and indoor/outdoor imaging.
Beveridge et al.~\cite{Beveridge2009} analyzed results for three algorithms on a Face Recognition Grand Challenge~\cite{frgc} dataset, and found that males had a higher verification rate for the two higher-accuracy algorithms and females had a higher verification rate for the third, lower-accuracy algorithm. 
Klare et al~\cite{Klare2012} presented results from three commercial off-the-shelf (COTS) and three research algorithms showing that females had a worse receiver operating characteristic (ROC) curve for all six, and also showed example results for which females had both a worse impostor distribution and a worse genuine distribution.
Grother et al.~\cite{Grother2010} analyzed the Multiple Biometric Evaluation (MBE) results and found that women have a higher false non-match rate (FNMR) 
at a fixed false match rate for 6 of the 7 matchers evaluated. However, they did not investigate the relationship between impostor scores and gender.
Ueda et al.~\cite{ueda2010influence} report that face recognition is advantaged by the presence of light makeup and disadvantaged by the presence of heavy makeup. 
The authors conclude that light makeup enhances the distinctiveness of faces by moderate changes in the facial features, whereas heavy makeup increases the symmetry in a way that reduces the distinctiveness.

\subsection{Fairness in Face Recognition After Deep Learning}

The above works are from before the deep learning wave in face recognition. 
Related works since the rise of deep learning matchers report similar results. 

Cook et al.~\cite{cook2018} analyze images acquired using eleven different automated kiosks with the same COTS matcher, and report that genuine scores are lower for females.
Lu et al.~\cite{Lu2018} use deep CNN matchers and datasets from the IARPA Janus program in a detailed study of the effect of various covariates, and report that accuracy is lower for females.
Howard et al.~\cite{bio_rally} report higher false match rates for African-American and Caucasian females under age 40. For subjects over 40, African-American females again had a higher false match rate, but Caucasian females had a lower false match rate than Caucasian males. However, the dataset is the smallest of the recent papers on this topic, with 363 subjects divided into 8 subsets for this analysis.
Vera-Rodriguez et al.~\cite{Vera-Rodriguez2019} use a Resnet-50 and a VGGFace matcher with the VGGFace2 dataset~\cite{vggface2}, and report that females have a worse ROC for both.
Albiero et al.~\cite{Albiero2020_gender} use the ArcFace matcher with four datasets,
and report that the general pattern of results is that females have a worse ROC curve, impostor distribution and genuine distribution.
Similarly, Krishnapriya et al.~\cite{KrishnapriyaTTS} show that both the ArcFace and the VGGFace2 matcher, which are trained on different datasets and with different loss functions, result in a worse ROC curve, impostor distribution and genuine distribution for females, for both African-American and Caucasian image cohorts.
Grother et al.~\cite{frvt3}, in the 2019 FRVT focused on demographic analysis, found that females have higher false match rates (worse impostor distribution), and that the phenomenon is consistent across a wide range of different matchers and datasets. 
Relative to the genuine distribution, they report that women often have higher false negative error rates, but note that there are exceptions to this generalization.

The general pattern across these works, for a broad variety of matchers and datasets, is that face recognition accuracy is lower for females, with females having a worse impostor distribution and a worse genuine distribution.

Several researchers have speculated makeup use as a cause of the gender gap in accuracy~\cite{Klare2012, cook2018, Lu2018}.
Dantcheva et al.~\cite{can_facial_cosmetics} and others have documented how the use of makeup can degrade face recognition accuracy.
In scenarios where females show substantial makeup use and males do not, this is a possible cause of accuracy difference.
However, Albiero et al.~\cite{Albiero2020_gender} showed that makeup use plays a minor role in the MORPH dataset, which is a main dataset used in this paper.
Albiero et al.~\cite{Albiero2020_gender} also concluded that hairstyles are not a driving factor in accuracy difference, but they considered hair only in the context of occluding the forehead.
In this paper, we consider hair occlusion as it affects skin visibility over the whole face.
Another speculated cause of accuracy difference is shorter average height for women, leading to non-optimal camera angle~\cite{Albiero2020_gender, Grother2010, cook2018}.

Few works attempt to identify the cause(s) of the accuracy gender gap.
For example, the recent NIST report on demographic effects~\cite{frvt3} lists ``analyze cause and effect’’ as an item under ``what we did not do’’.
However, Albiero et al.~\cite{Albiero2020_gender} reported on experiments to determine if makeup use, facial expression, forehead occlusion by hair, or downward-looking camera angle could explain the worse impostor and genuine distributions observed for females.
They found that these factors, either individually or together, did not significantly impact the gender gap in accuracy.
A popular speculation is that accuracy is lower for females because they are under-represented in the training data. Albiero et al.~\cite{Albiero2020_train} trained matchers from scratch using male / female ratios of 0/100, 25/75, 50/50, 75/25 and 100/0.
They found that explicitly gender-balanced training data does not result in balanced accuracy on test data. They also found that a 25\% male / 75\% female training set resulted in the least-imbalanced accuracy on the test data, 
but was generally not the training mix that maximized female, male, or average accuracy.

\begin{figure*}[t]
  \begin{subfigure}[b]{1\linewidth}
      \begin{subfigure}[b]{0.32\linewidth}
        \centering
          \begin{subfigure}[b]{0.23\columnwidth}
            \centering
            \includegraphics[width=\linewidth]{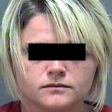}
          \end{subfigure}
          \begin{subfigure}[b]{0.23\columnwidth}
            \centering
            \includegraphics[width=\linewidth]{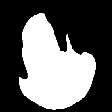}
          \end{subfigure}
          \begin{subfigure}[b]{0.23\columnwidth}
            \centering
            \includegraphics[width=\linewidth]{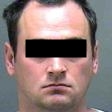}
          \end{subfigure}
          \begin{subfigure}[b]{0.23\columnwidth}
            \centering
            \includegraphics[width=\linewidth]{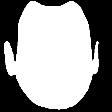}
          \end{subfigure}
          \caption{MORPH Caucasian}
          \vspace{-0.5em}
      \end{subfigure}
      \hfill
      \begin{subfigure}[b]{0.32\linewidth}
        \centering
          \begin{subfigure}[b]{0.23\columnwidth}
            \centering
            \includegraphics[width=\linewidth]{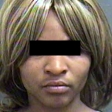}
          \end{subfigure}
          \begin{subfigure}[b]{0.23\columnwidth}
            \centering
            \includegraphics[width=\linewidth]{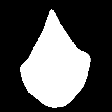}
          \end{subfigure}
          \begin{subfigure}[b]{0.23\columnwidth}
            \centering
            \includegraphics[width=\linewidth]{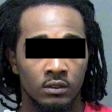}
          \end{subfigure}
          \begin{subfigure}[b]{0.23\columnwidth}
            \centering
            \includegraphics[width=\linewidth]{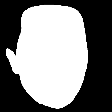}
          \end{subfigure}
          \caption{MORPH African-American}
          \vspace{-0.5em}
      \end{subfigure}
      \hfill
        \begin{subfigure}[b]{0.32\linewidth}
          \centering
          \begin{subfigure}[b]{0.23\columnwidth}
            \centering
            \includegraphics[width=\linewidth]{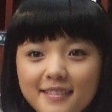}
          \end{subfigure}
          \begin{subfigure}[b]{0.23\columnwidth}
            \centering
            \includegraphics[width=\linewidth]{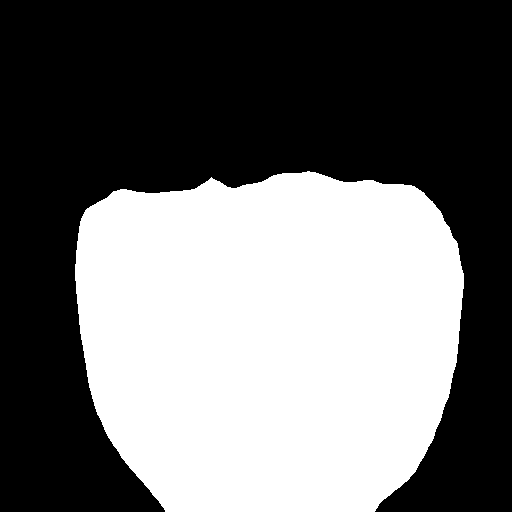}
          \end{subfigure}
          \begin{subfigure}[b]{0.23\columnwidth}
            \centering
            \includegraphics[width=\linewidth]{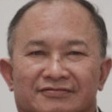}
          \end{subfigure}
          \begin{subfigure}[b]{0.23\columnwidth}
            \centering
            \includegraphics[width=\linewidth]{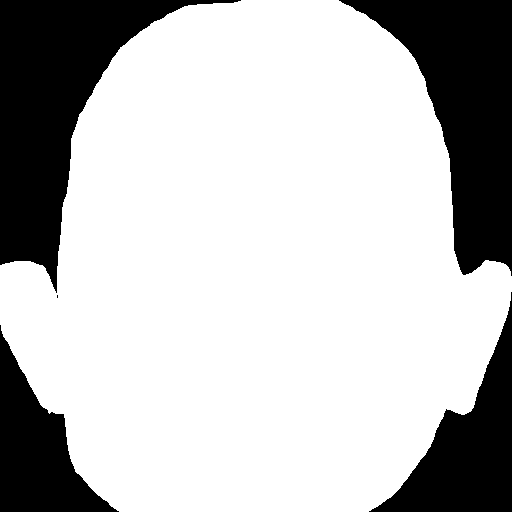}
          \end{subfigure}
          \caption{Asian-Celeb}
          \vspace{-0.5em}
      \end{subfigure}
  \end{subfigure}
  \caption{Example images and their face/non-face masks based on BiSeNet segmentation (eye regions of MORPH images blacked out for this figure as a privacy consideration).}
  \vspace{-0.5em}
  \label{fig:samples}
\end{figure*}

\subsection{Gender Classification from Face}
Gender classification of a face image is a related problem but distinct from face recognition.
Buolamwini and Gebru~\cite{gendershades}, Muthukumar et al.~\cite{Muthukumar} and others have looked at the accuracy of algorithms that predict gender from a face image, and found lower accuracy for females.
Balakrishnan et al.~\cite{balakrishnan2020towards} reports on a method for measuring algorithmic bias in gender prediction, where the authors manipulate the face attributes in an image, e.g. gender.
Their experimental results reveal bias caused by gender, hair length, age, and facial hair, but not by skin color.
Overall, past works report that gender-from-face classification has a similar pattern to face recognition, with females having lower accuracy than males, where a similar cause (hairstyle) is one of the main issues.
Qiu et al.~\cite{Qiu_2021} investigate the relationship between errors in gender classification and errors in face recognition.
They find that an image that results in a gender classification error is {\it less} likely to participate in a false match error in face recognition, but more likely to participate in a false non-match error.

\section{Datasets}

The MORPH dataset~\cite{MORPH, MORPH_site} was originally collected to support research in face aging, and has been widely used in that context.
It has also recently been used in the study of demographic variation in accuracy~\cite{Albiero2020_gender, Albiero2020_train, KrishnapriyaTTS, albiero2019does, krishnapria_cvprw_2019}.
MORPH contains mugshot-style images that are nominally frontal pose, neutral expression and acquired with controlled lighting and an 18\% gray background.
We curated the MORPH 3 dataset in order to remove duplicate images, twins, and mislabeled images for
35,276 images of 8,835 Caucasian males,
10,941 images of 2,798 Caucasian females,
56,245 images of 8,839 African-American males,
and 24,857 images of 5,929 African-American females.
Sample images from the different MORPH demographic sets are shown in Figure~\ref{fig:samples}.

The second dataset selected is the Asian-Celeb dataset~\cite{asian_celeb}.
Asian-Celeb is web-scraped dataset, mainly composed of Asian faces, containing a range of illumination, pose and quality variation.
In order for this dataset to be more comparable to MORPH, we curated it following the steps suggested by~\cite{curation_method}, which included removal of mislabeled images and duplicate images, and pose constraint.
Then, using a race classifier from~\cite{Albiero2020_train}, we removed images classified as not Asian. 
After curation, we ended with 73,376 images of 12,673 Asian males, and 43,356 images of 6,083 Asian females.
Sample images from Asian-Celeb are shown in Figure~\ref{fig:samples}.
(The list of image IDs for this curated subset of AsianCeleb will be made publicly available.)

We chose the MORPH dataset as well-suited for this research because it was collected in a controlled environment, with similar illumination, pose and resolution across images, reducing the possibility of accidental differences between female and male images that could influence accuracy analysis.
Since to the best of our knowledge, there is no public Asian faces dataset of significant size that is collected in a controlled environment, we started with the largest in-the-wild Asian dataset we could find, Asian-Celeb, and curated it to get a subset more comparable to the image quality in MORPH. Having both datasets with as few uncontrolled factors as practical should allow a more accurate inference about the cause(s) of observed differences. As shown in Figure~\ref{fig:makeup_dist}, makeup plays a significantly larger role in Asian-Celeb than in MORPH. 
This likely results from the fact that in-the-wild, web-scraped datasets are mostly images of ``celebrities’’ appearing in public, and celebrities in public making greater use of cosmetics.

\section{Matchers}

For our experiments, we used three matchers: two open-source, ArcFace and a gender-balanced-training CNN matcher; and one commercial (COTS).
The instance of ArcFace~\cite{arcface} used here corresponds to a set of publicly-available weights~\cite{insightface}, trained on the MS1MV2 dataset, which is a publicly-available, ``cleaned'' version of MS1M~\cite{ms1_celeb}.
The gender-balanced-training matcher used was released by~\cite{Albiero2020_train}. It was explicitly trained using a gender-balanced version of MS1MV2 dataset, where the same amount of subjects and images for both males and females are used.
It was trained using a combined margin loss (ArcFace margin~\cite{arcface}, CosFace margin~\cite{cosface}, and SphereFace margin~\cite{sphereface}). 
For more training details, see~\cite{albiero2019does}.
The commercial-off-the-shelf (COTS) matcher is a recent (2020) version, among the top-ranking matchers according to NIST~\cite{frvt3}. For license reasons, we cannot disclose the COTS name.
The input to both matchers are aligned faces resized to 112x112 (ArcFace, gender-balanced matcher) and 224x224 (COTS) using~\cite{albiero2021img2pose, retinaface}.
For ArcFace, 512-d features are extracted, and matched using cosine similarity.

\section{Experimental Results}
\subsection{Lower Recognition Accuracy For Females}

\begin{figure*}[t]
  \begin{subfigure}[b]{1\linewidth}
      \begin{subfigure}[b]{0.32\linewidth}
        \centering
          \includegraphics[width=\linewidth]{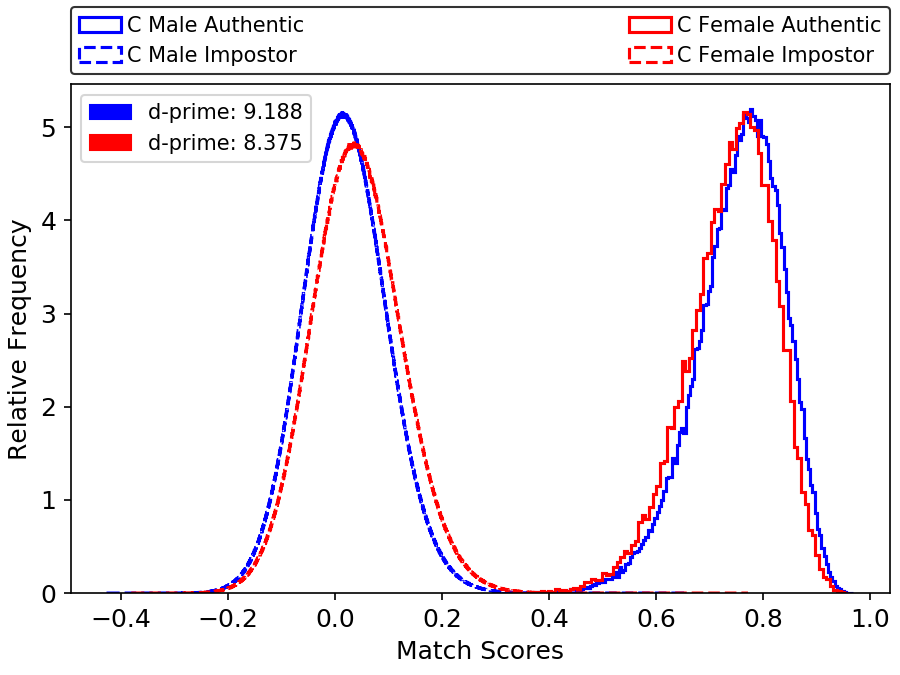}
      \end{subfigure}
      \hfill 
      \begin{subfigure}[b]{0.32\linewidth}
        \centering
          \includegraphics[width=\linewidth]{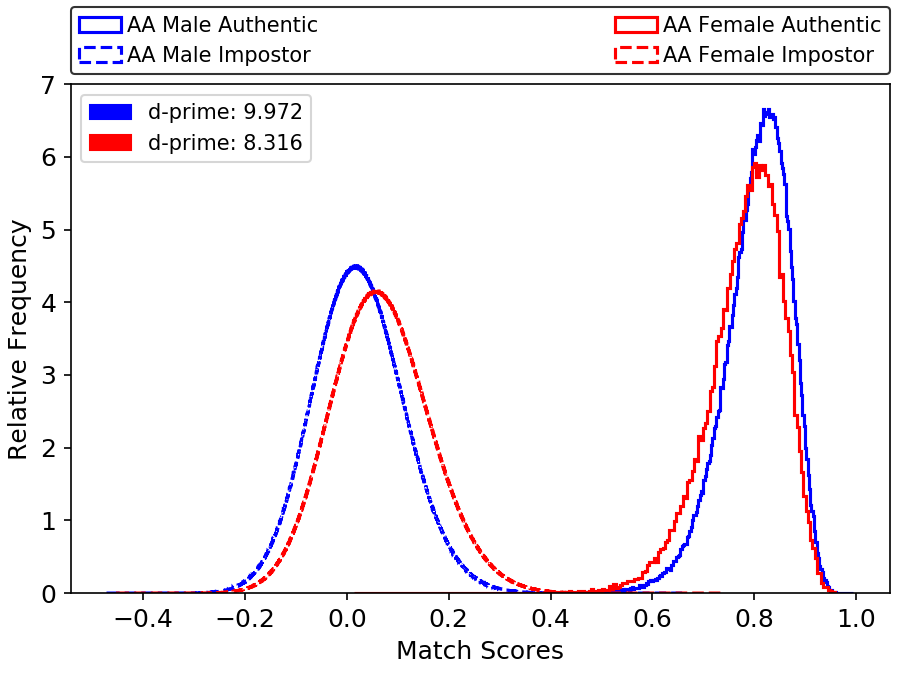}
      \end{subfigure}
      \hfill 
      \begin{subfigure}[b]{0.32\linewidth}
        \centering
          \includegraphics[width=\linewidth]{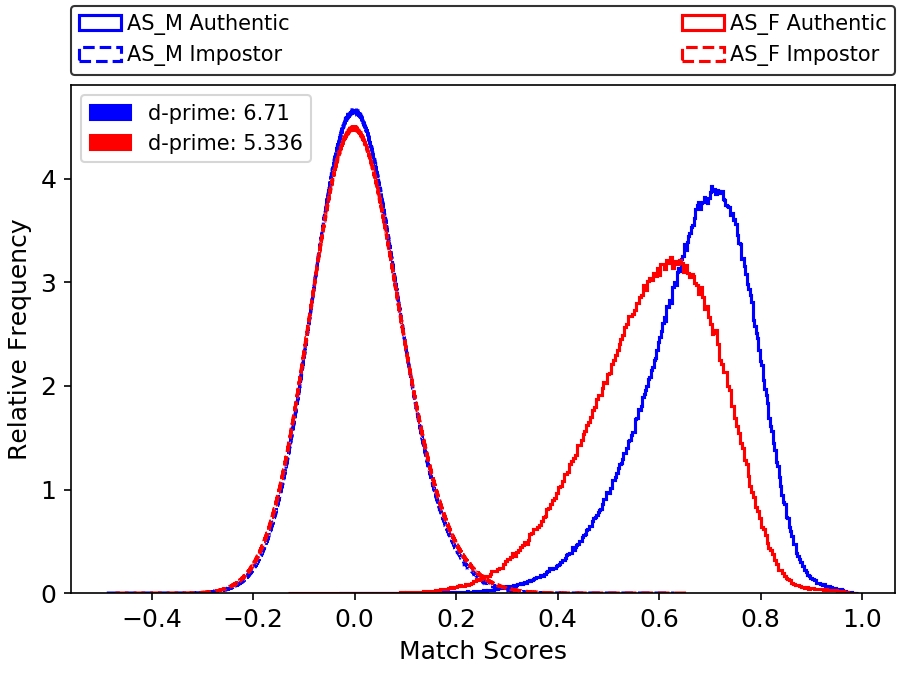}
      \end{subfigure}
  \end{subfigure}
  
  \begin{subfigure}[b]{1\linewidth}
      \begin{subfigure}[b]{0.32\linewidth}
        \centering
          \includegraphics[width=\linewidth]{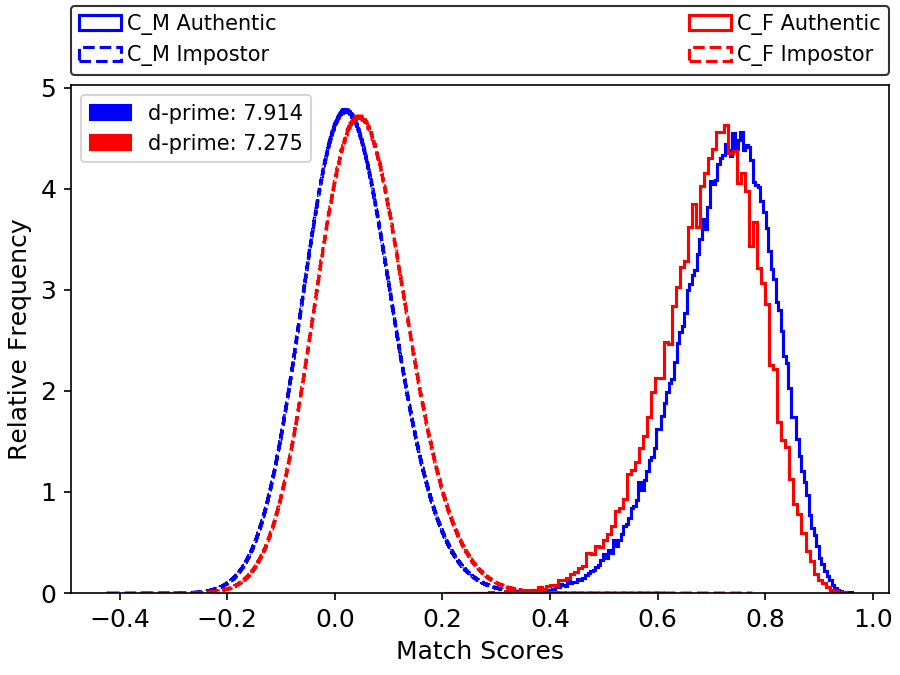}
      \end{subfigure}
      \hfill 
      \begin{subfigure}[b]{0.32\linewidth}
        \centering
          \includegraphics[width=\linewidth]{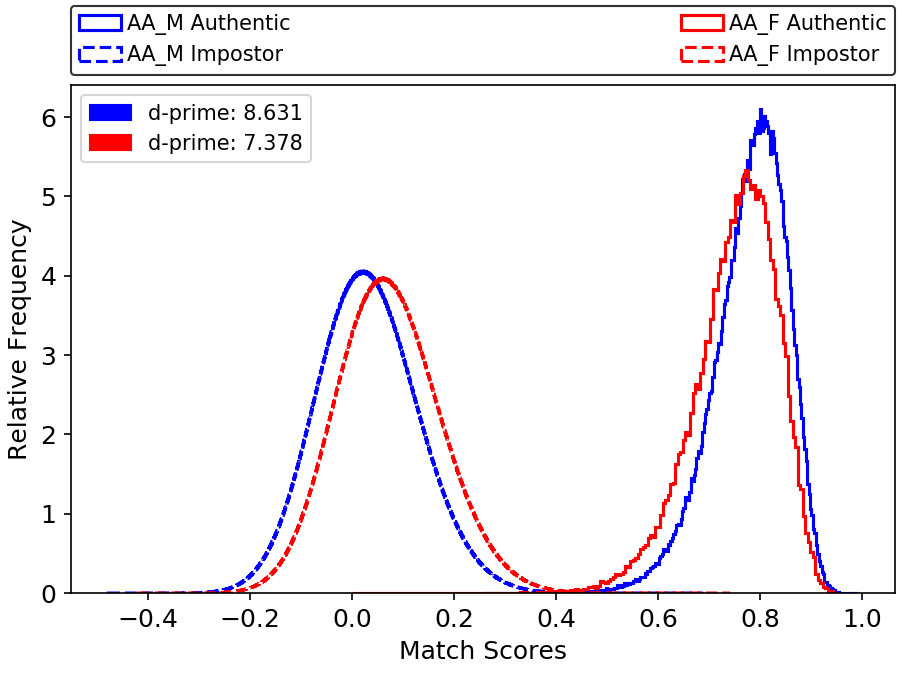}
      \end{subfigure}
      \hfill 
      \begin{subfigure}[b]{0.32\linewidth}
        \centering
          \includegraphics[width=\linewidth]{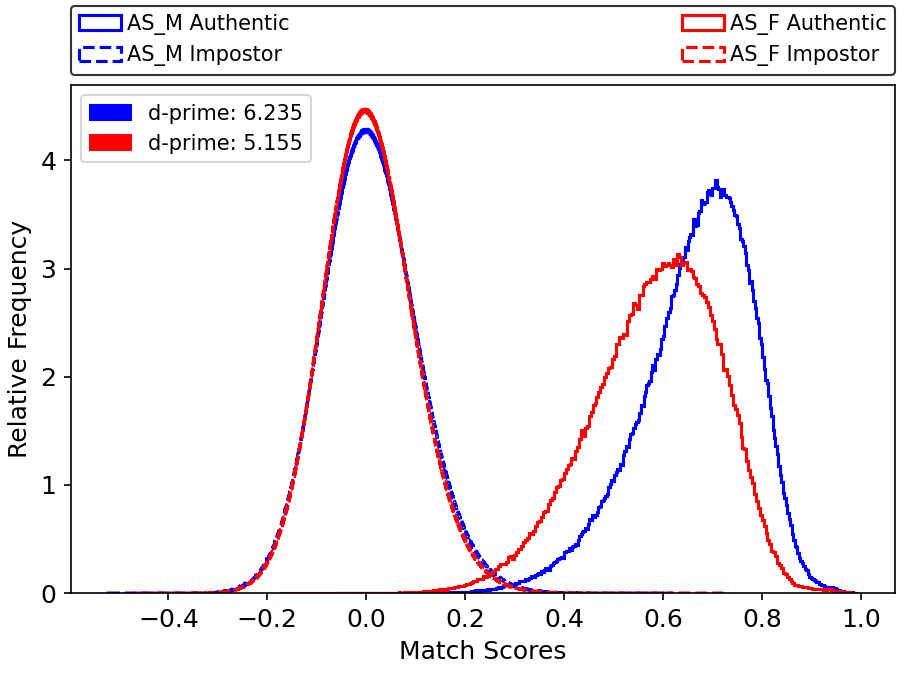}
      \end{subfigure}
  \end{subfigure}
  
  \begin{subfigure}[b]{1\linewidth}
      \begin{subfigure}[b]{0.32\linewidth}
        \centering
          \includegraphics[width=\linewidth]{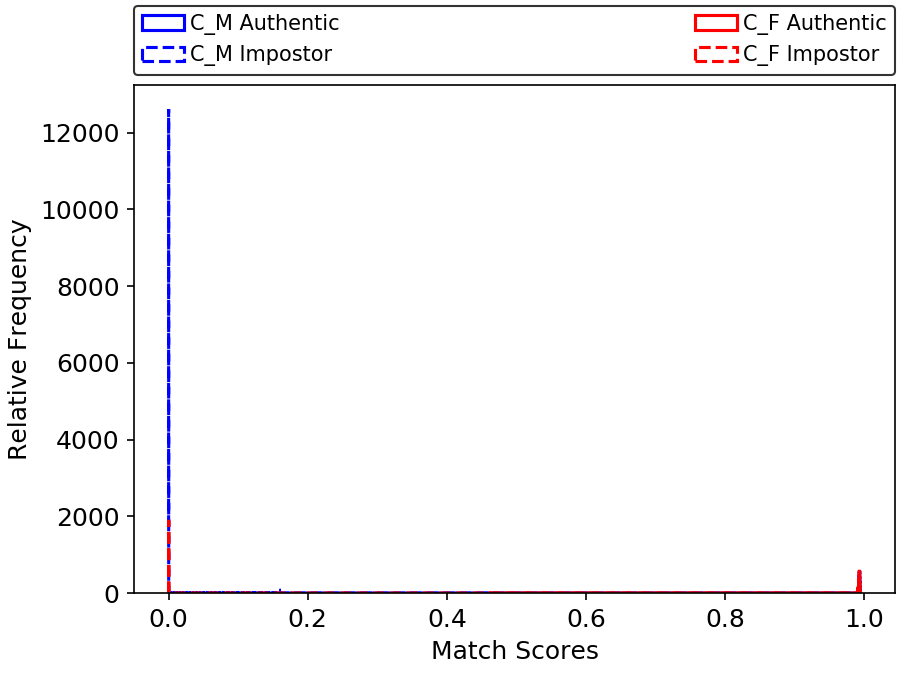}
          \caption{MORPH Caucasian}
      \end{subfigure}
      \hfill 
      \begin{subfigure}[b]{0.32\linewidth}
        \centering
          \includegraphics[width=\linewidth]{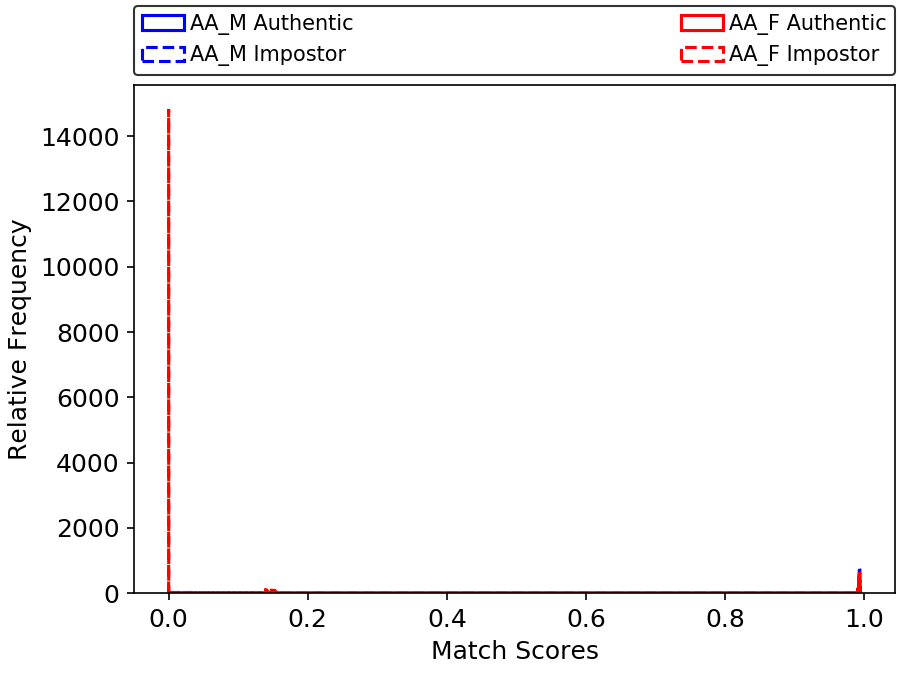}
          \caption{MORPH African-American}
      \end{subfigure}
      \hfill 
      \begin{subfigure}[b]{0.32\linewidth}
        \centering
          \includegraphics[width=\linewidth]{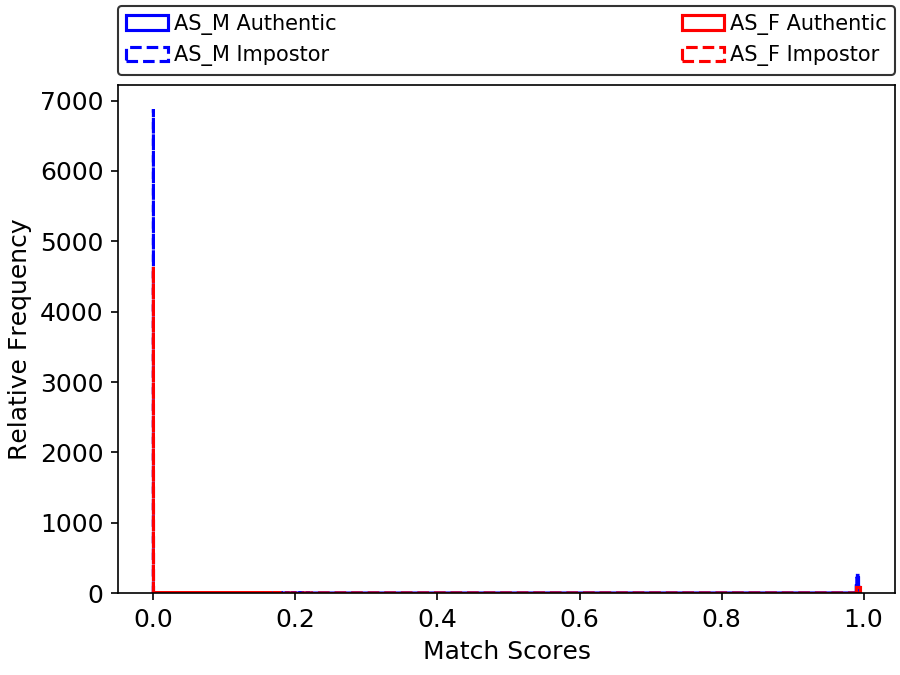}
          \caption{Asian-Celeb}
      \end{subfigure}
        
  \end{subfigure}
  \caption{Impostor and genuine distributions for ArcFace (top), gender-balanced matcher (middle), and COTS (bottom) matcher. For ArcFace and balanced-gender-training matcher, the female impostor distribution ranges over higher similarity scores (except for Asian-Celeb), indicating a higher FMR, and the genuine distribution ranges over lower similarity scores, indicating a higher FNMR. For COTS, a significant portion of the values for each distribution is mapped to either 0 or 1, so that neither distribution gives a typical appearance.  This reflects an unknown vendor-specific mapping of the raw similarity scores, as is a ``feature'' of analyzing results from a COTS matcher.}
  \vspace{-1.0em}
  \label{fig:auth_imp}
\end{figure*}

The lower accuracy for females can be quantified in terms of
the separation between the female and male impostor and genuine distributions. For the ArcFace matcher and gender-balanced matcher, as shown in the top of Figure~\ref{fig:auth_imp}. 
With the exception of Asian-Celeb with the gender-balanced matcher, for all three datasets, the female impostor distribution is shifted toward higher similarity scores than the male impostor distribution.
Also for all three datasets, the female genuine distribution is shifted toward lower similarity scores, with the result that the impostor-to-genuine separation is lower for females.
For the COTS matcher, as the distributions are distinctly not Gaussian, we cannot make the same comparison (see bottom of Figure~\ref{fig:auth_imp}).
For this reason, we move our analysis from genuine and impostor distribution to false match rate (FMR) and false non-match rate (FNMR).
Both FMR and FNMR were computed for each group separately.

\begin{figure*}[t]
  \begin{subfigure}[b]{1\linewidth}
      \begin{subfigure}[b]{0.32\linewidth}
        \centering
          \includegraphics[width=\linewidth]{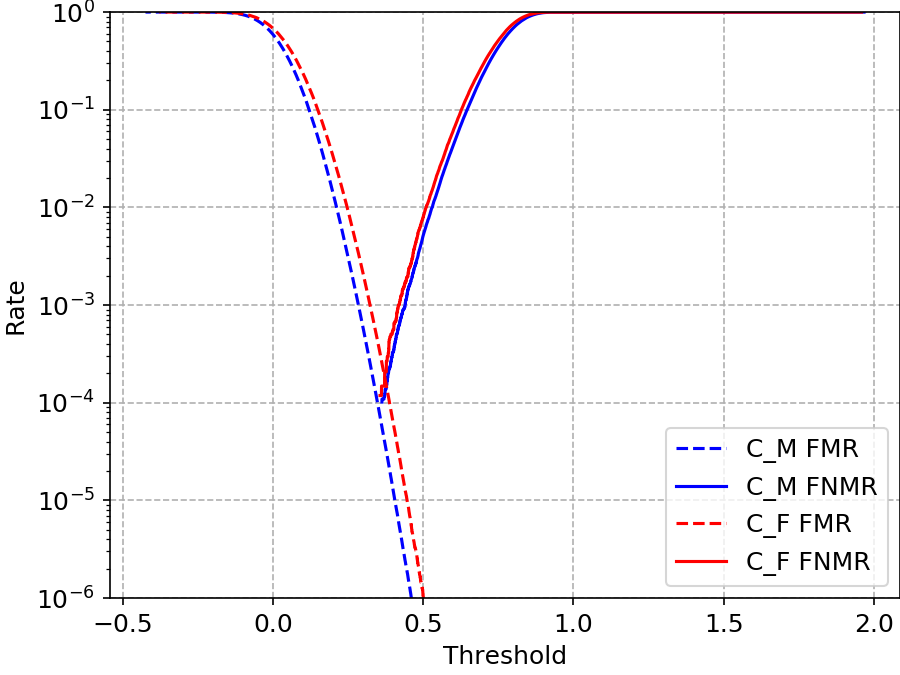}
      \end{subfigure}
      \hfill 
      \begin{subfigure}[b]{0.32\linewidth}
        \centering
          \includegraphics[width=\linewidth]{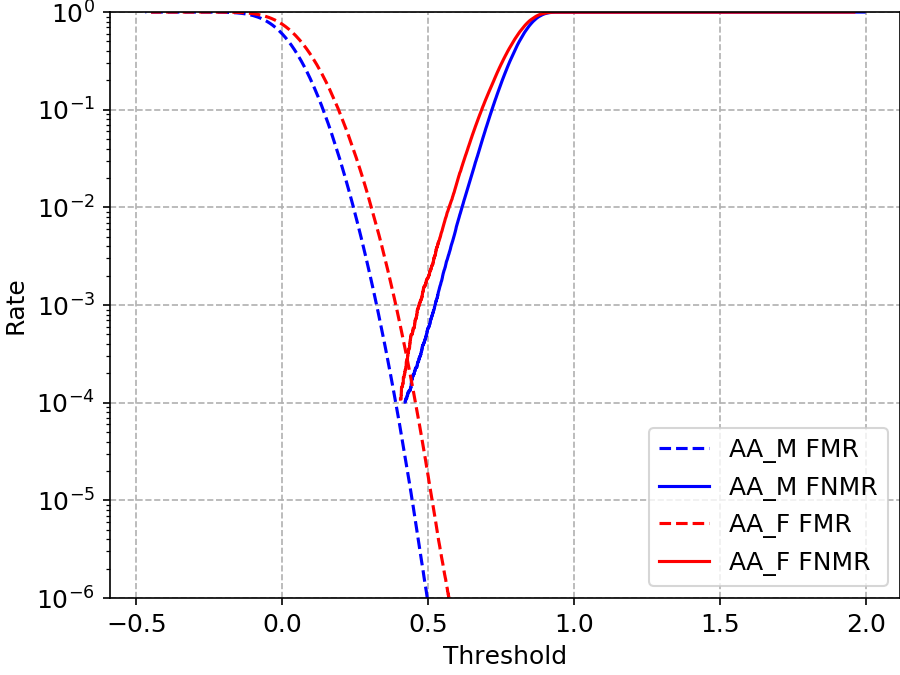}
      \end{subfigure}
      \hfill 
      \begin{subfigure}[b]{0.32\linewidth}
        \centering
          \includegraphics[width=\linewidth]{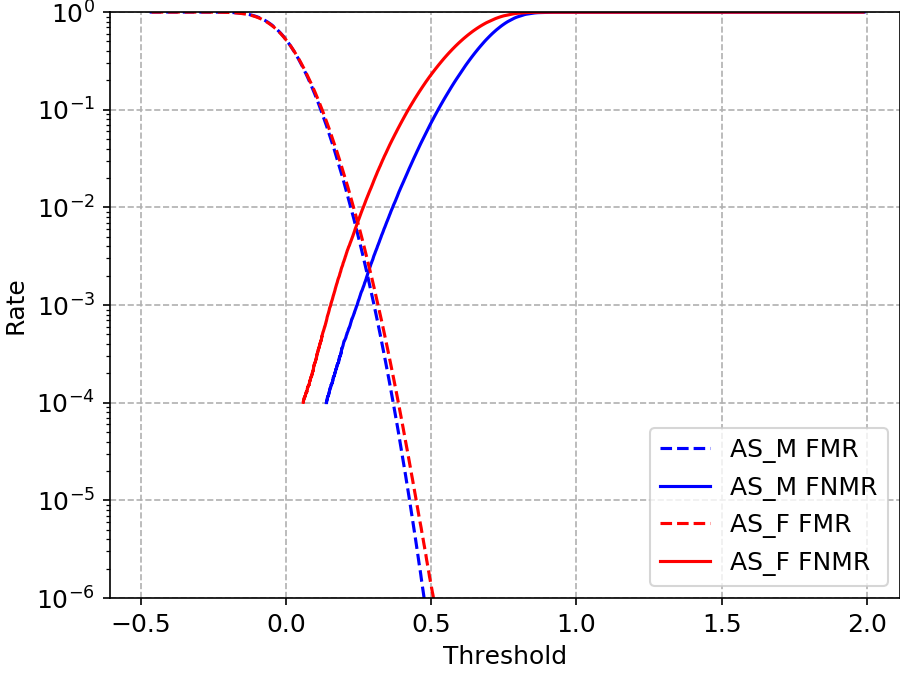}
      \end{subfigure}
  \end{subfigure}
  
  \begin{subfigure}[b]{1\linewidth}
      \begin{subfigure}[b]{0.32\linewidth}
        \centering
          \includegraphics[width=\linewidth]{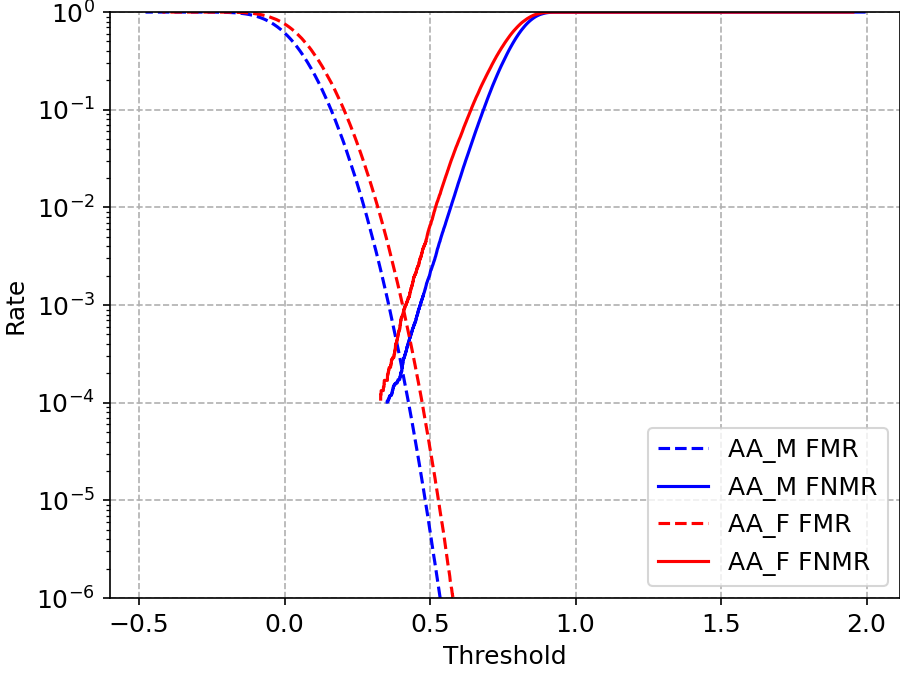}
      \end{subfigure}
      \hfill 
      \begin{subfigure}[b]{0.32\linewidth}
        \centering
          \includegraphics[width=\linewidth]{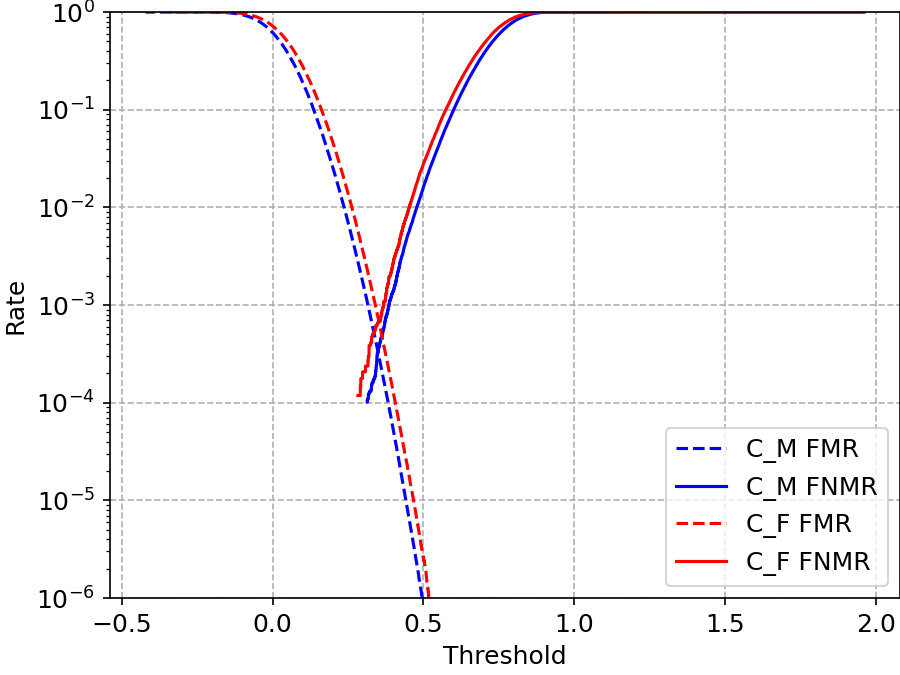}
      \end{subfigure}
      \hfill 
      \begin{subfigure}[b]{0.32\linewidth}
        \centering
          \includegraphics[width=\linewidth]{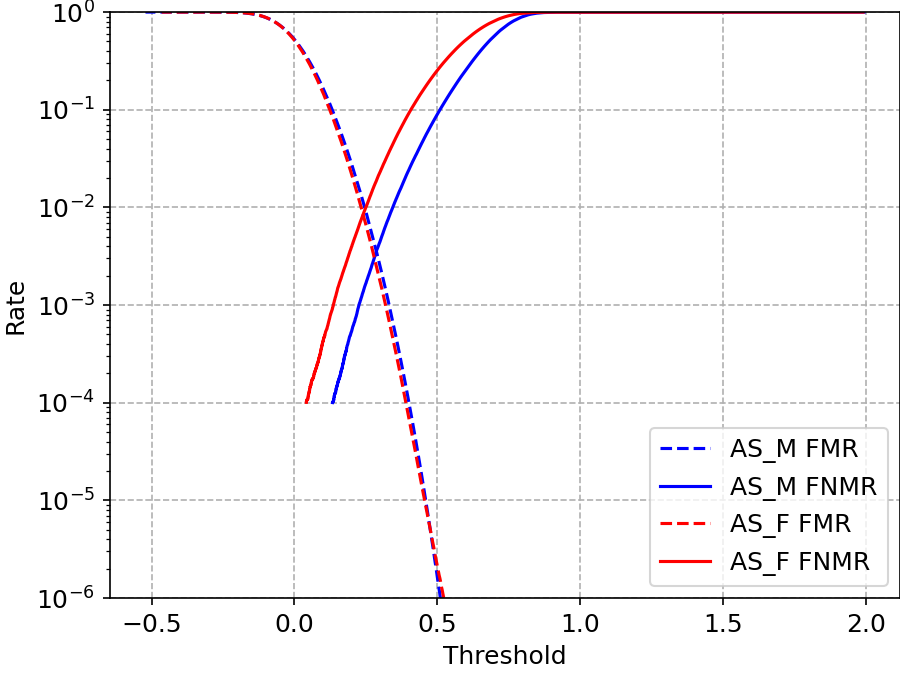}
      \end{subfigure}
  \end{subfigure}
  \begin{subfigure}[b]{1\linewidth}
      \begin{subfigure}[b]{0.32\linewidth}
        \centering
          \includegraphics[width=\linewidth]{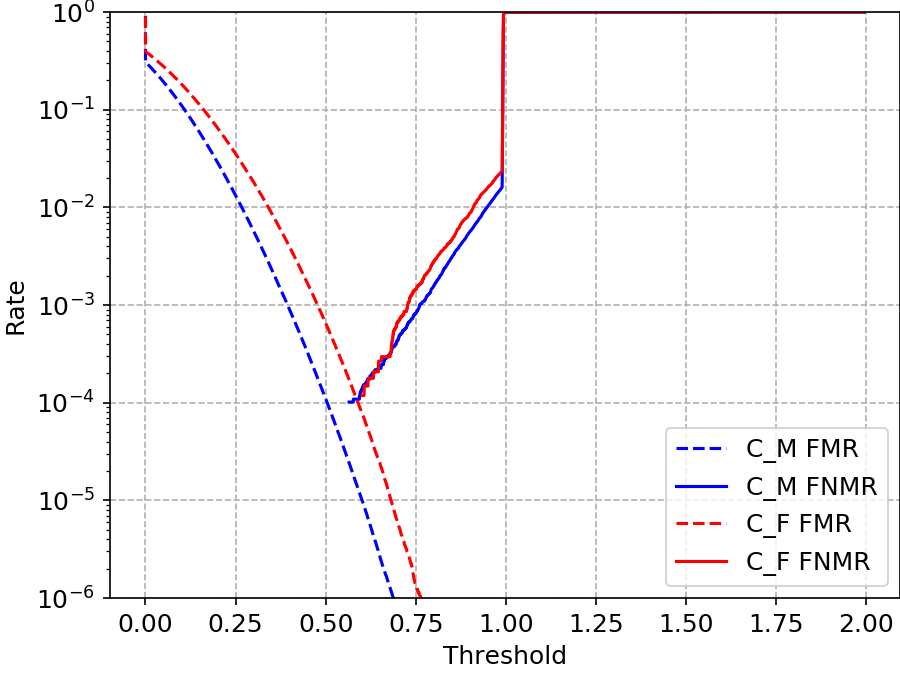}
          \caption{MORPH Caucasian}
      \end{subfigure}
      \hfill 
      \begin{subfigure}[b]{0.32\linewidth}
        \centering
          \includegraphics[width=\linewidth]{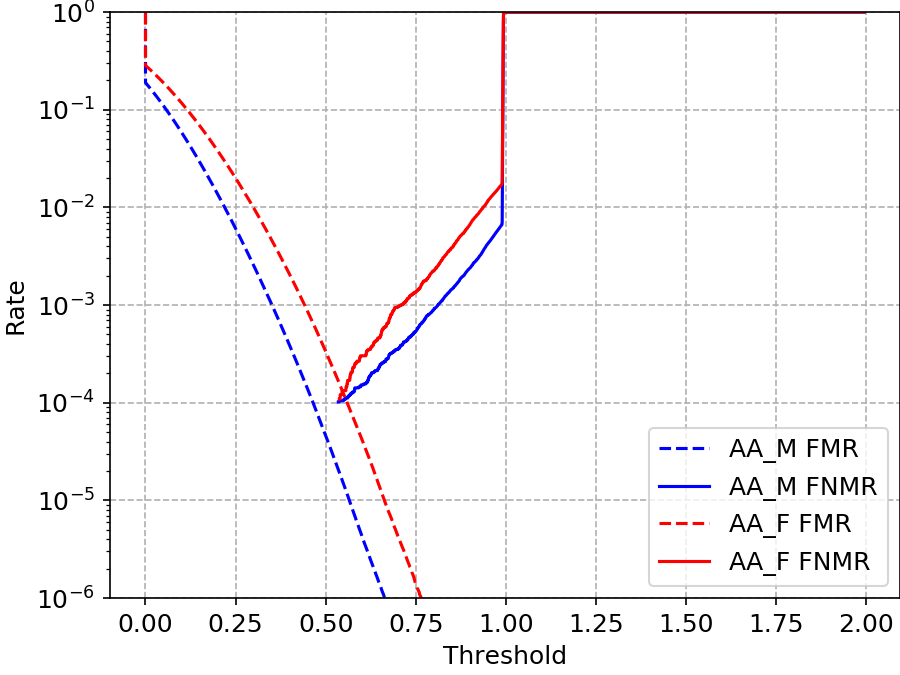}
          \caption{MORPH African-American}
      \end{subfigure}
      \hfill 
      \begin{subfigure}[b]{0.32\linewidth}
        \centering
          \includegraphics[width=\linewidth]{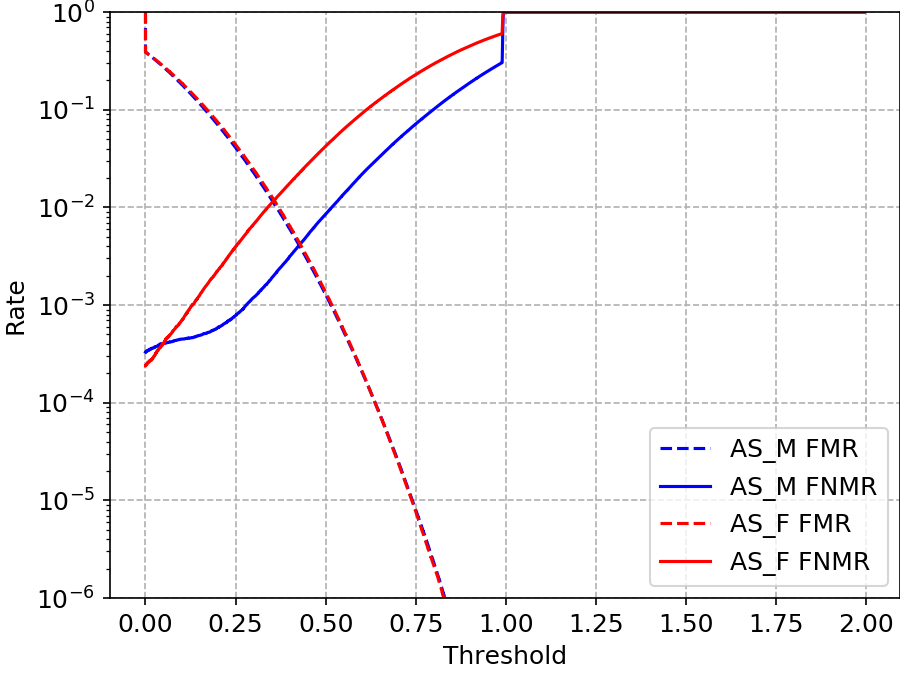}
          \caption{Asian-Celeb}
      \end{subfigure}
  \end{subfigure}
  \caption{FMR and FNMR for ArcFace (top), gender-balanced matcher (middle), and COTS (bottom) matchers; females have higher FMR for all but one case, and higher and FNMR for all cases.}
  \vspace{-1.0em}
  \label{fig:fmr_fnmr}
\end{figure*}

FMR and FNMR for both matchers are shown in Figure~\ref{fig:fmr_fnmr}. 
Here we can see that both FMR and FNMR are generally worse for females for both matchers across all datasets.
For both MORPH datasets, FMR is much higher for females than males. However, for Asian-Celeb, they are more similar.
We speculate that this difference can be attributed to the fact that Asian-Celeb is a web-scraped dataset, and so the impostor pairs might still, even after curation, have factors other than gender affecting the FMR.
The FNMR is consistently higher across all datasets, whereas on Asian-Celeb and MOPRH African-American the difference is larger than MORPH Caucasian.

\subsection{Less Pixels-On-Face In Female Images}

This section considers the fraction of the image that is input to matchers
that represents ``pixels on the face'', and how this differs between female and male.
The faces in the MORPH images are detected and aligned, so that faces are centered in a standard location that is the input to the matchers.
We used the Asian-Celeb images as provided in~\cite{asian_celeb}, as the images were provided with faces already aligned, cropped and resized to 112x112 pixels.
Examples are shown in Figure~\ref{fig:samples}. 

To obtain a binary mask that indicates which pixels represent face and which do not, we use Bilateral Segmentation Network (``BiSeNet'')~\cite{bisenet} to segment the faces.
A pre-trained version of BiSeNet~\cite{bisenet_github} segments face images into regions.
For our purposes, ``face'' is the union of BiSeNet regions 1 to 13, corresponding to skin, eyebrows, eyes, ears, nose and mouth. 

\begin{figure*}[t]
  \begin{subfigure}[b]{1\linewidth}
    \begin{subfigure}[b]{0.32\linewidth}
      \begin{subfigure}[b]{0.48\columnwidth}
        \centering
          \begin{subfigure}[b]{1\columnwidth}
            \centering
            \includegraphics[width=\linewidth]{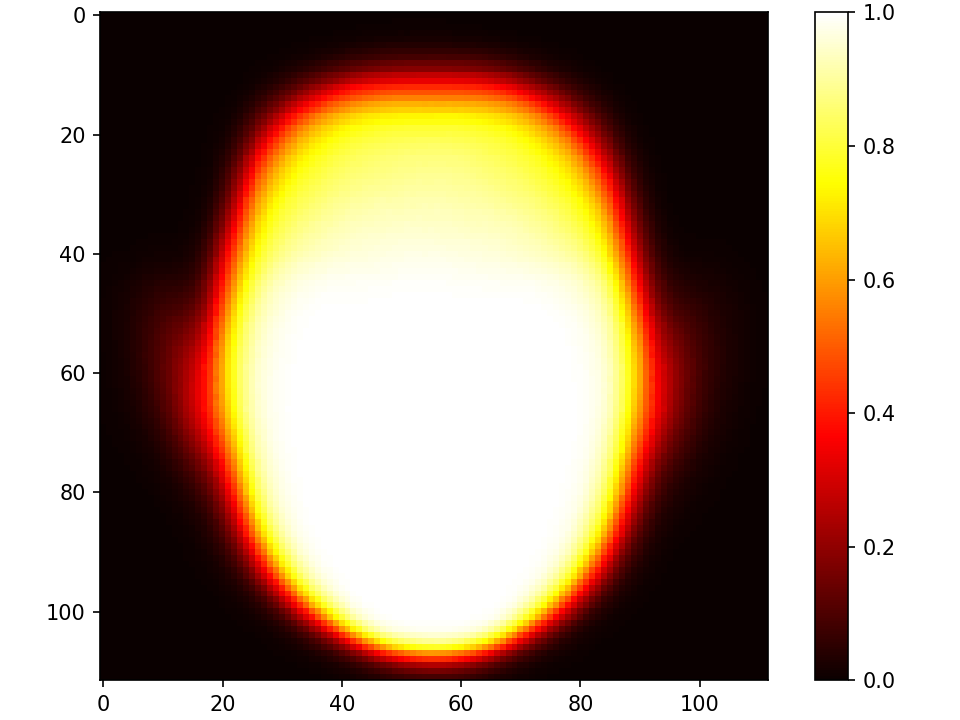}
          \end{subfigure}
          \begin{subfigure}[b]{1\columnwidth}
            \centering
            \includegraphics[width=\linewidth]{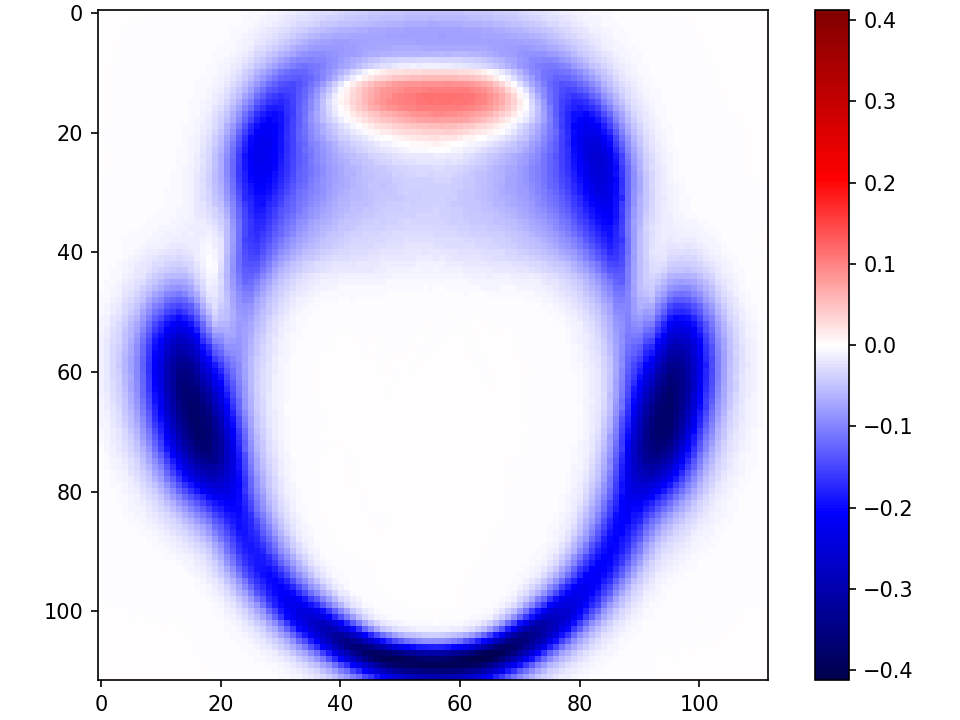}
          \end{subfigure}
          \begin{subfigure}[b]{1\columnwidth}
            \centering
            \includegraphics[width=\linewidth]{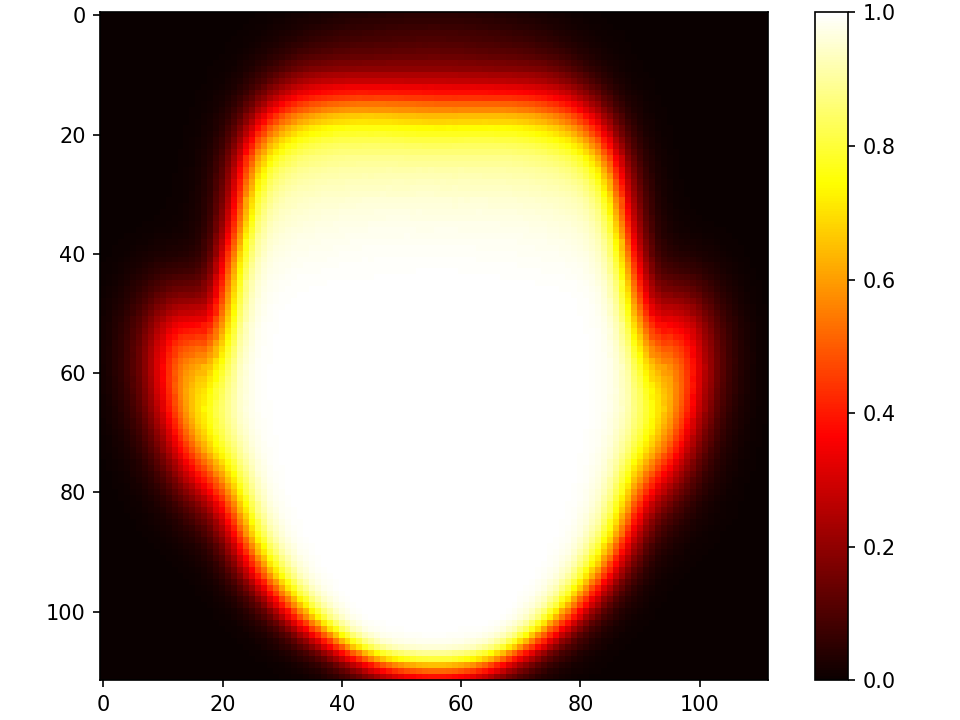}
          \end{subfigure}
      \end{subfigure}
      \begin{subfigure}[b]{0.48\columnwidth}
        \centering
          \begin{subfigure}[b]{1\columnwidth}
            \centering
            \includegraphics[width=\linewidth]{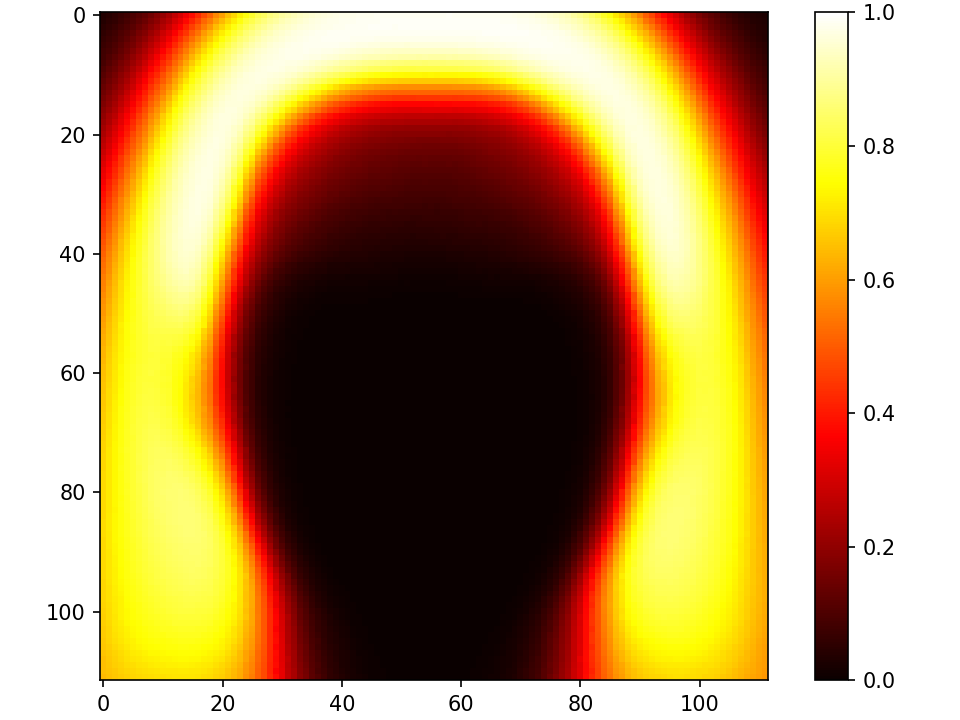}
          \end{subfigure}
          \begin{subfigure}[b]{1\columnwidth}
            \centering
            \includegraphics[width=\linewidth]{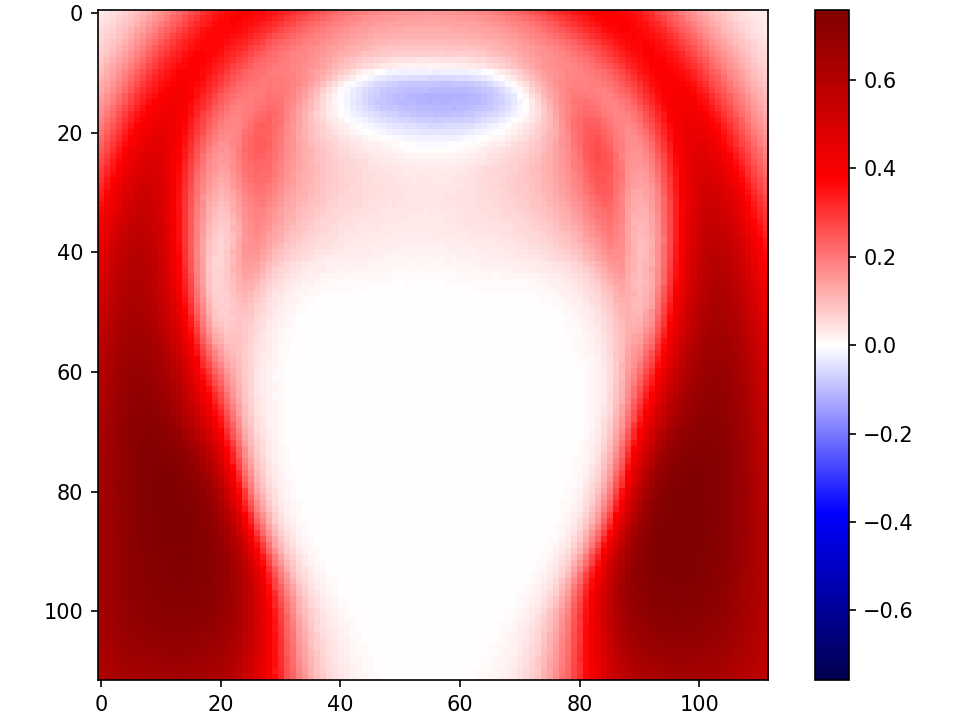}
          \end{subfigure}
          \begin{subfigure}[b]{1\columnwidth}
            \centering
            \includegraphics[width=\linewidth]{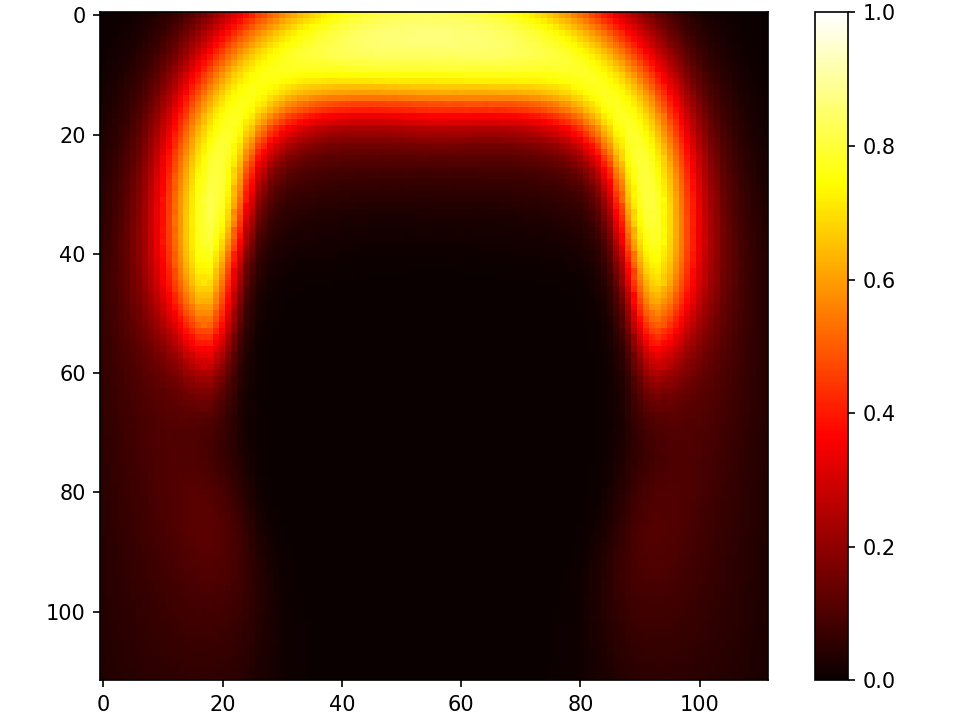}
          \end{subfigure}
        \end{subfigure}
          \caption{MORPH Caucasian}
      \end{subfigure}
      \hfill 
      \begin{subfigure}[b]{0.32\linewidth}
      \begin{subfigure}[b]{0.48\columnwidth}
        \centering
          \begin{subfigure}[b]{1\columnwidth}
            \centering
            \includegraphics[width=\linewidth]{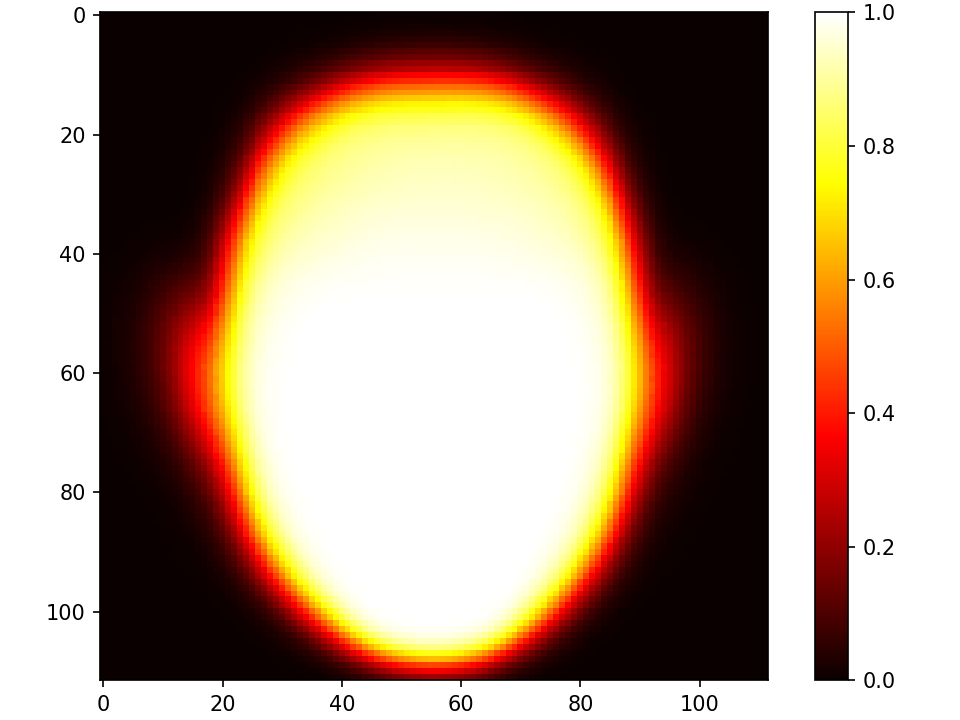}
          \end{subfigure}
          \begin{subfigure}[b]{1\columnwidth}
            \centering
            \includegraphics[width=\linewidth]{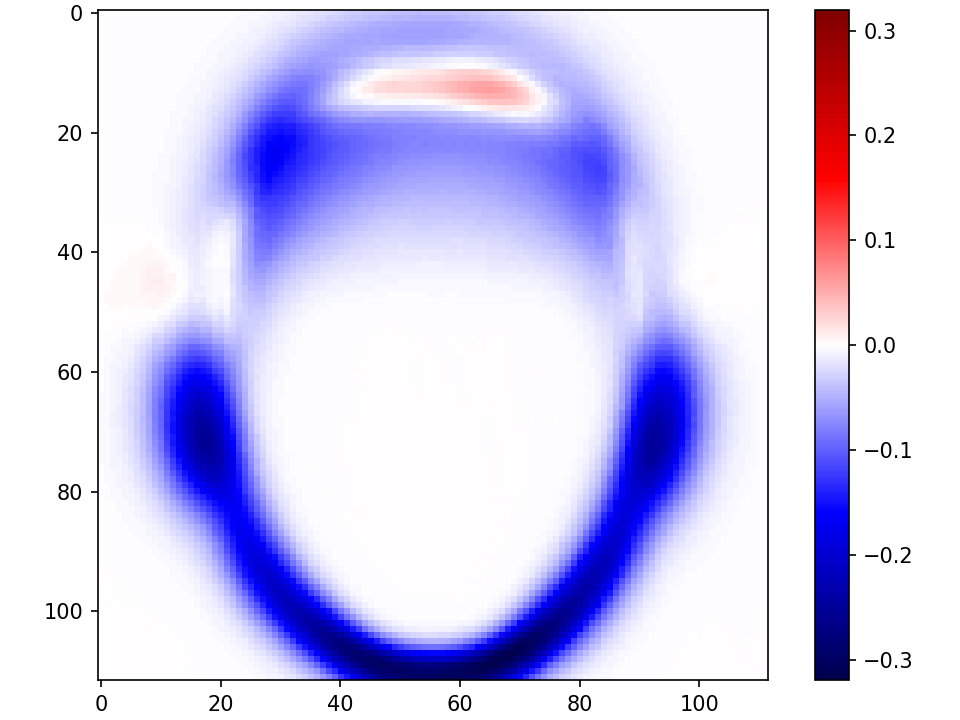}
          \end{subfigure}
          \begin{subfigure}[b]{1\columnwidth}
            \centering
            \includegraphics[width=\linewidth]{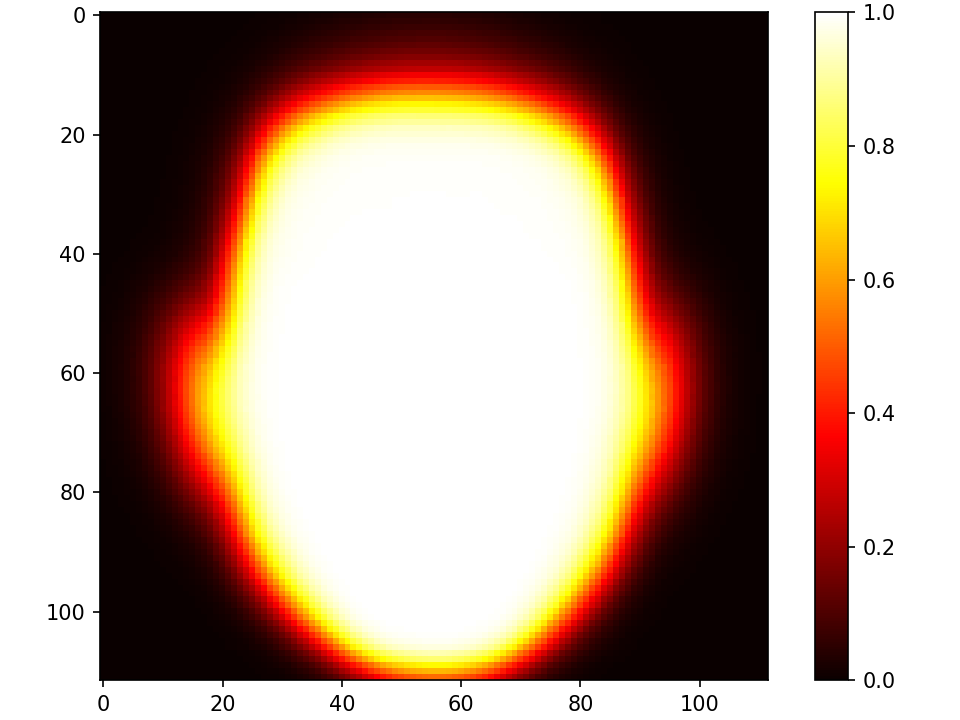}
          \end{subfigure}
      \end{subfigure}
          \begin{subfigure}[b]{0.48\columnwidth}
        \centering
          \begin{subfigure}[b]{1\columnwidth}
            \centering
            \includegraphics[width=\linewidth]{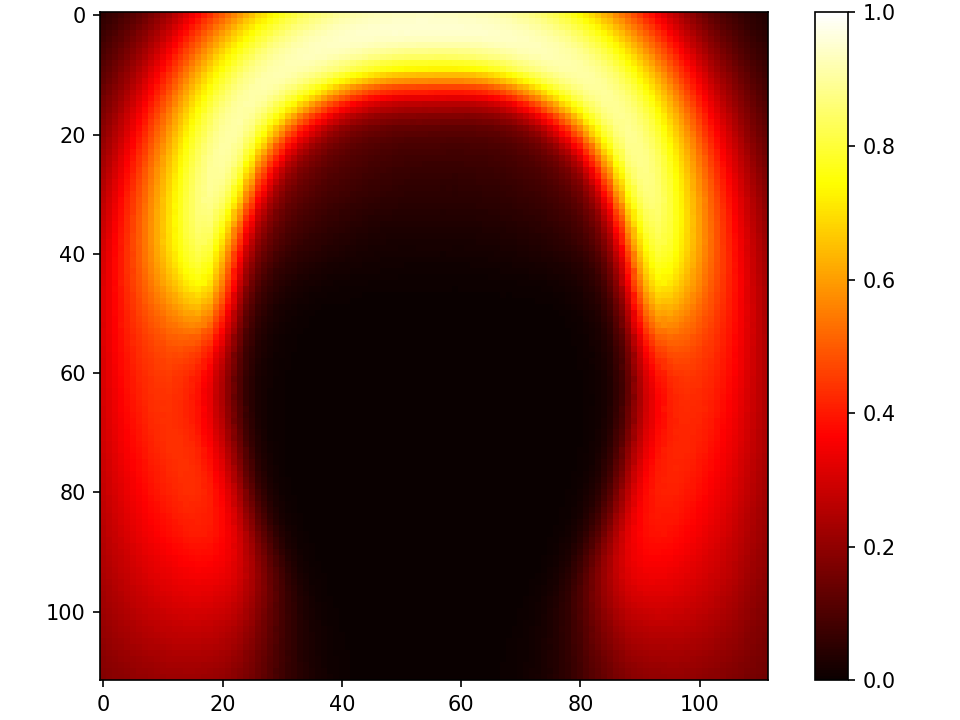}
          \end{subfigure}
          \begin{subfigure}[b]{1\columnwidth}
            \centering
            \includegraphics[width=\linewidth]{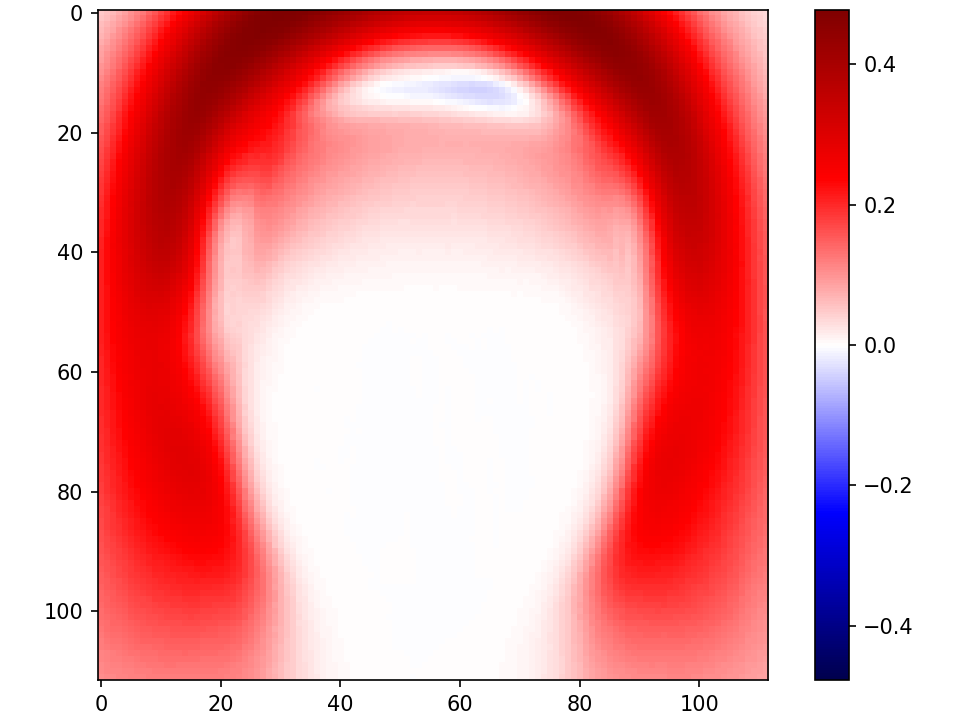}
          \end{subfigure}
          \begin{subfigure}[b]{1\columnwidth}
            \centering
            \includegraphics[width=\linewidth]{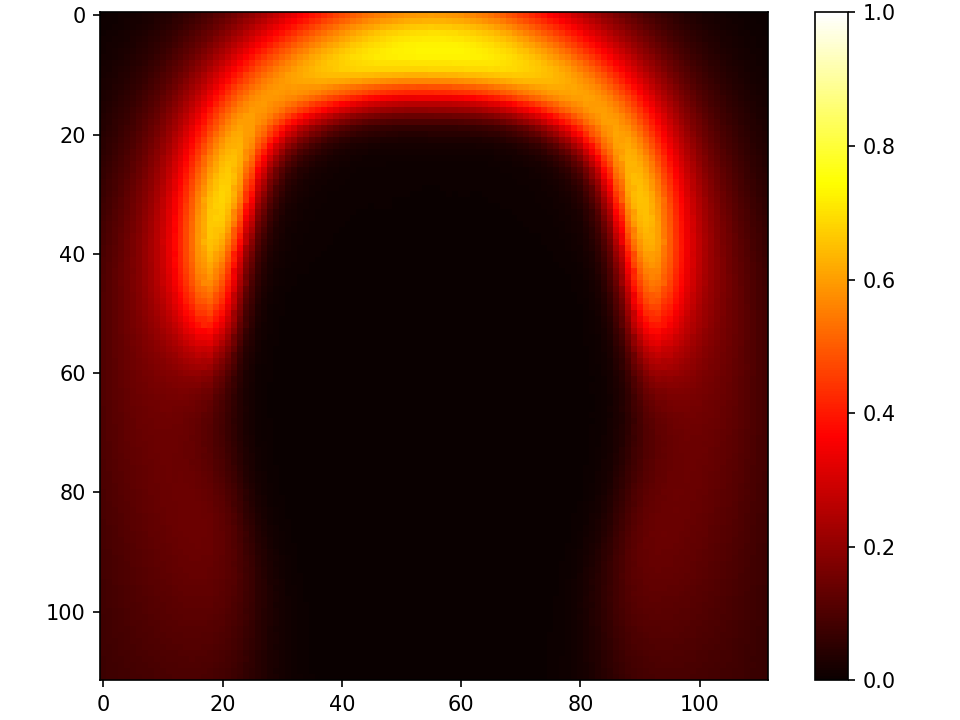}
          \end{subfigure}
        \end{subfigure}
          \caption{MORPH African-American}
      \end{subfigure}
      \hfill 
      \begin{subfigure}[b]{0.32\linewidth}
      \begin{subfigure}[b]{0.48\columnwidth}
        \centering
          \begin{subfigure}[b]{1\columnwidth}
            \centering
            \includegraphics[width=\linewidth]{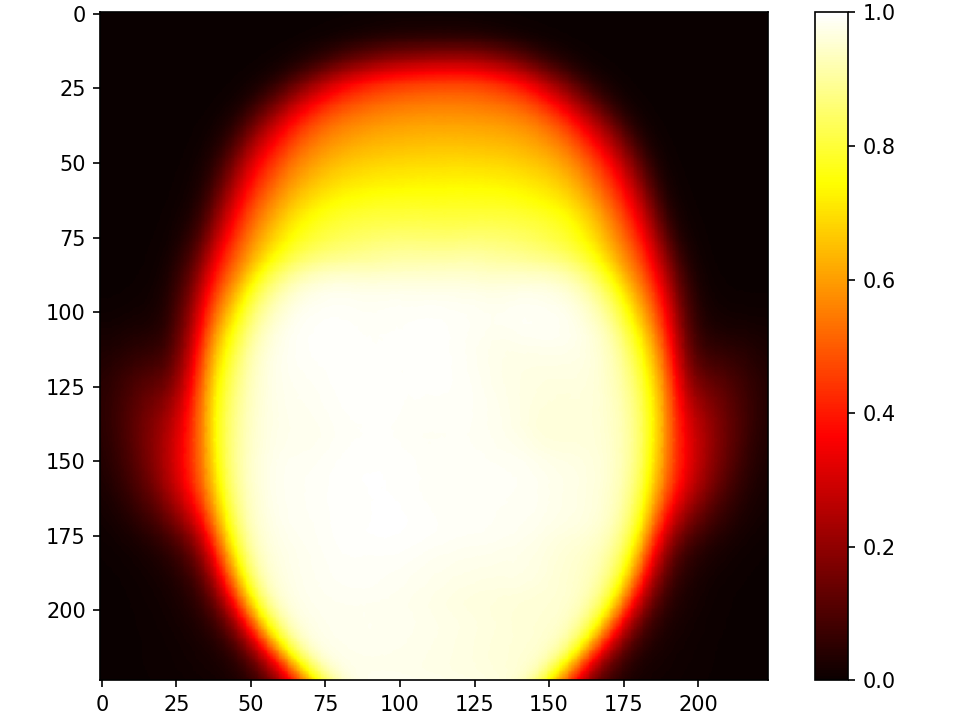}
          \end{subfigure}
          \begin{subfigure}[b]{1\columnwidth}
            \centering
            \includegraphics[width=\linewidth]{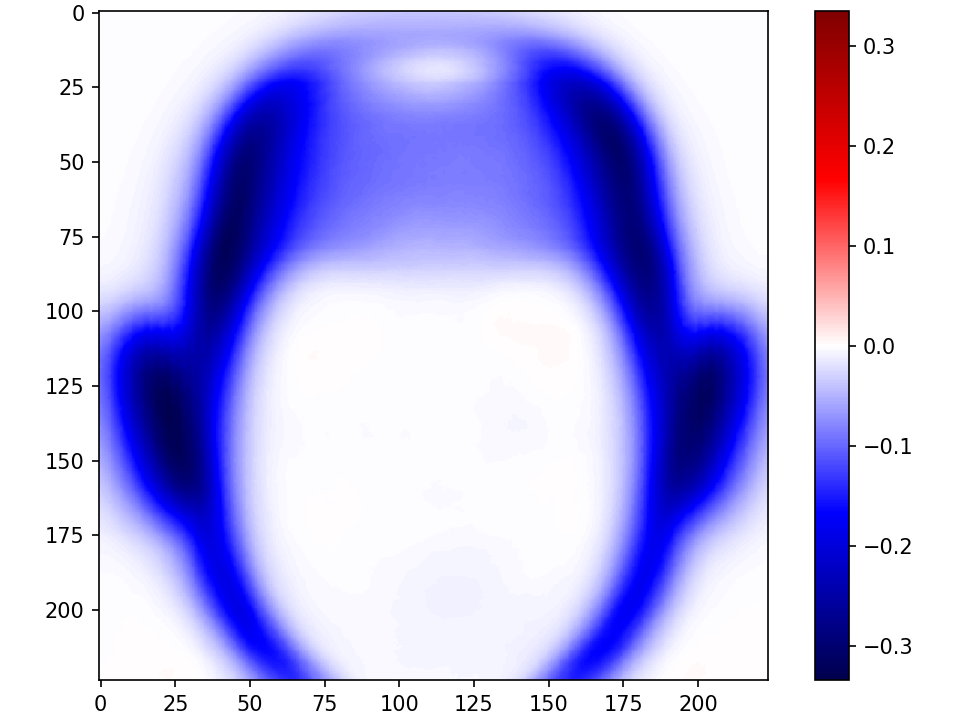}
          \end{subfigure}
          \begin{subfigure}[b]{1\columnwidth}
            \centering
            \includegraphics[width=\linewidth]{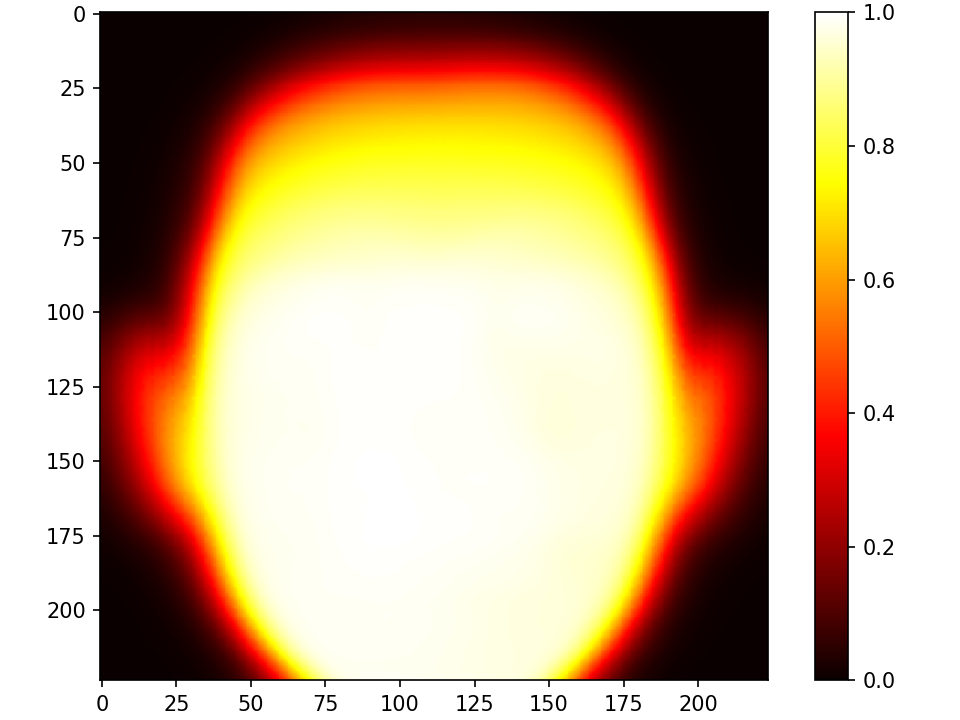}
          \end{subfigure}
      \end{subfigure}
      \begin{subfigure}[b]{0.48\columnwidth}
        \centering
          \begin{subfigure}[b]{1\columnwidth}
            \centering
            \includegraphics[width=\linewidth]{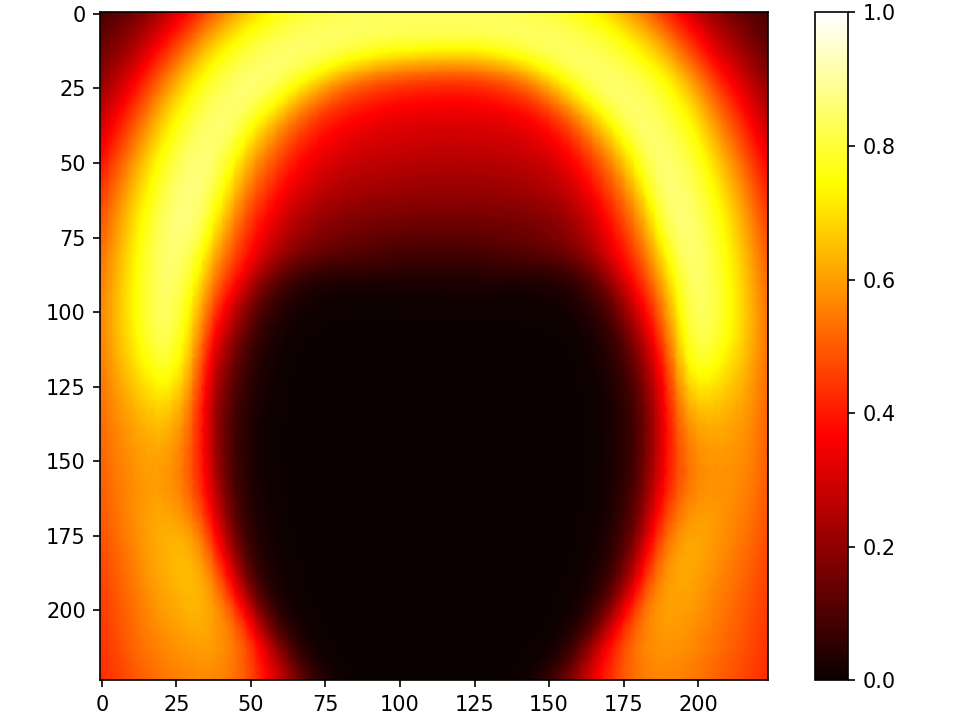}
          \end{subfigure}
          \begin{subfigure}[b]{1\columnwidth}
            \centering
            \includegraphics[width=\linewidth]{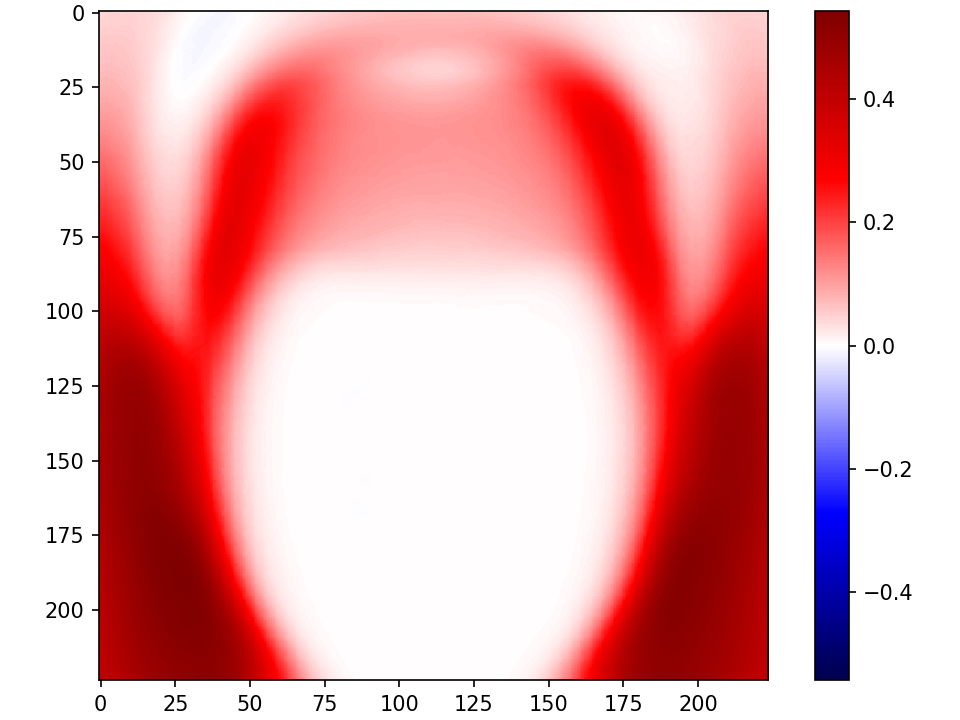}
          \end{subfigure}
          \begin{subfigure}[b]{1\columnwidth}
            \centering
            \includegraphics[width=\linewidth]{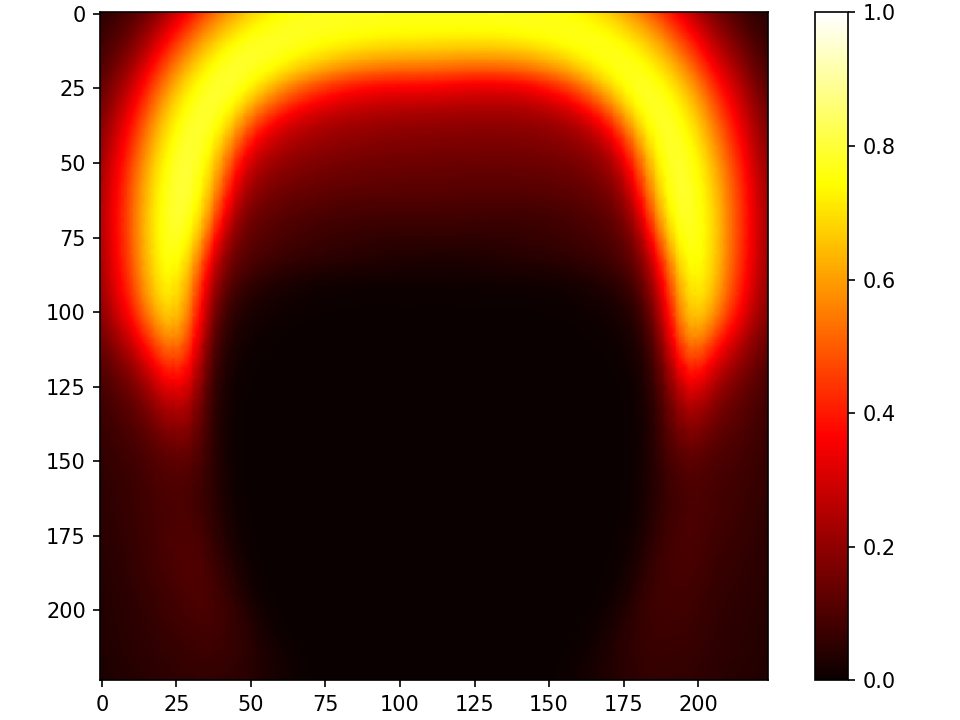}
          \end{subfigure}
         \end{subfigure}
          \caption{Asian-Celeb}
      \end{subfigure}
  \end{subfigure}
  \caption{Heatmaps representing frequency of pixels being labelled as ``face'' on left and as hair on right. Female heatmap is on top, male heatmap is on bottom, and male-female difference is on middle.}
  \vspace{-0.5em}
  \label{fig:heatmap}
\end{figure*}

Regions classified as neck, clothes, hair and hat are excluded.
Examples of the face/non-face masks are shown in Figure~\ref{fig:samples}.
These masks are used to compute heatmaps for female and male images. 
The value of each heatmap pixel ranges from 0 to 1, reflecting the fraction of images for which that pixel is labelled as face.
Figure~\ref{fig:heatmap} shows the female and male heatmaps and the difference between them.
The chin in the male heatmap extends slightly further toward the bottom.
This reflects the fact that females and males have, on average, different facial morphology; that is, different head size and shape~\cite{Farkas2005}.
Another difference is that the ear region and the sides of the face are less prominent in the female heatmap.
This reflects gender-associated hairstyles, where female hairstyles more frequently occlude the ears and part of the sides of the face.
The difference heatmaps summarize all of this, with blue representing pixels that more frequently labeled face for males than for females, and red representing the opposite.
The heatmaps make it clear that, {\it on average, female face images contain fewer ``pixels on the face'' than male face images, due to gendered hairstyles and different facial morphology.}

\begin{figure*}[t]
  \begin{subfigure}[b]{1\linewidth}
      \begin{subfigure}[b]{0.32\linewidth}
        \centering
          \includegraphics[width=\linewidth]{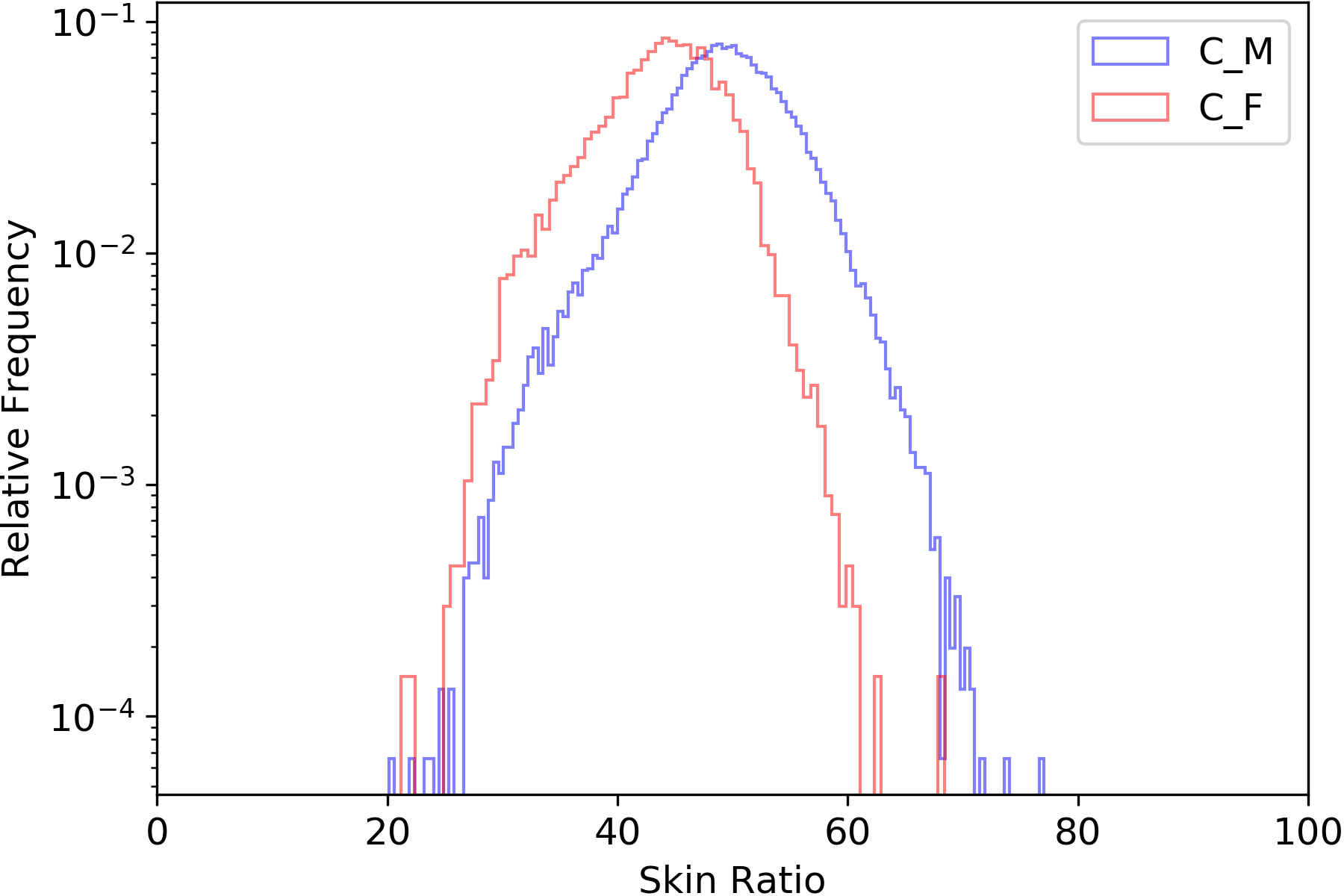}
          \caption{MORPH Caucasian}
          \vspace{-0.5em}
      \end{subfigure}
      \hfill 
      \begin{subfigure}[b]{0.32\linewidth}
        \centering
          \includegraphics[width=\linewidth]{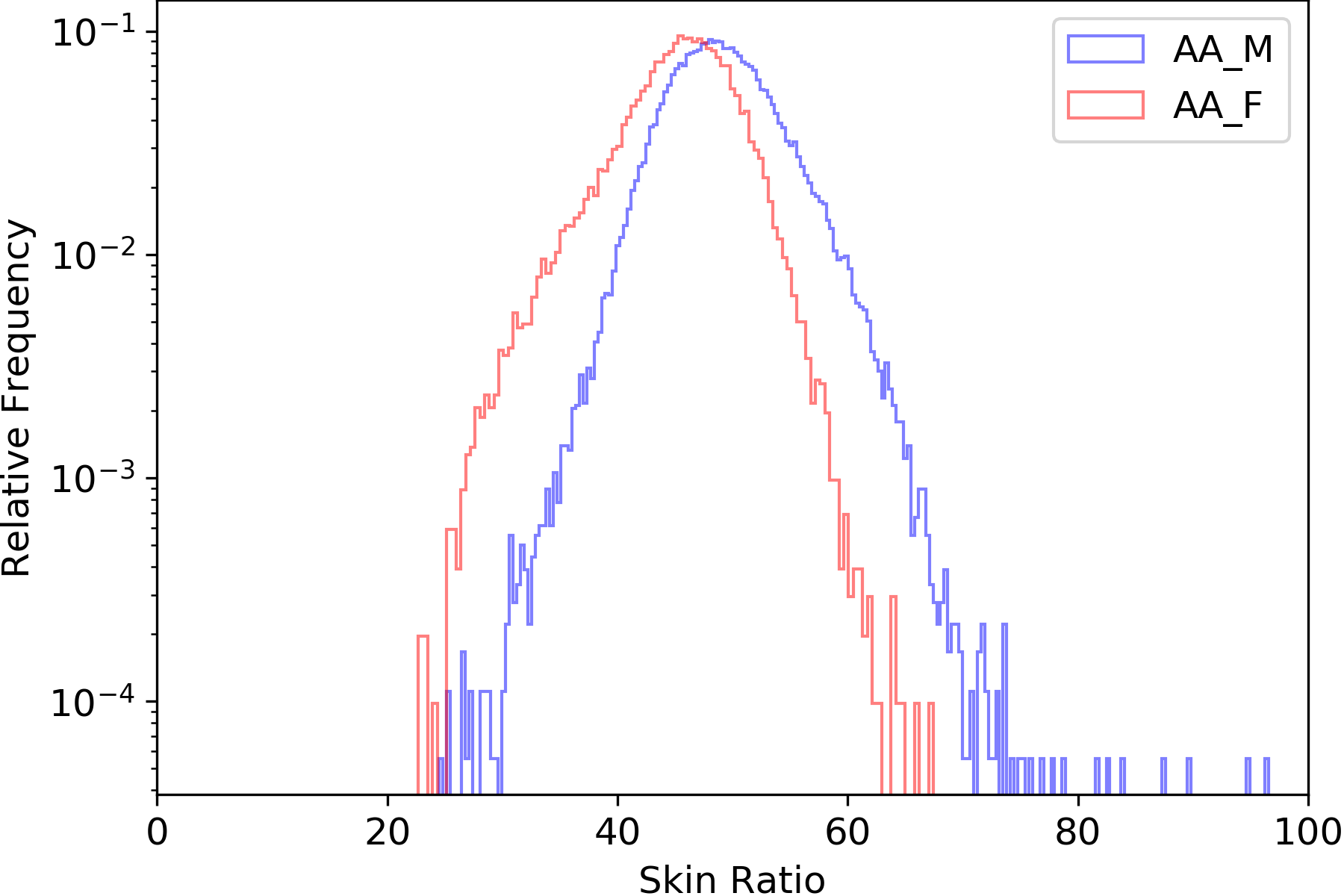}
          \caption{MORPH African-American}
          \vspace{-0.5em}
      \end{subfigure}
      \hfill 
      \begin{subfigure}[b]{0.32\linewidth}
        \centering
          \includegraphics[width=\linewidth]{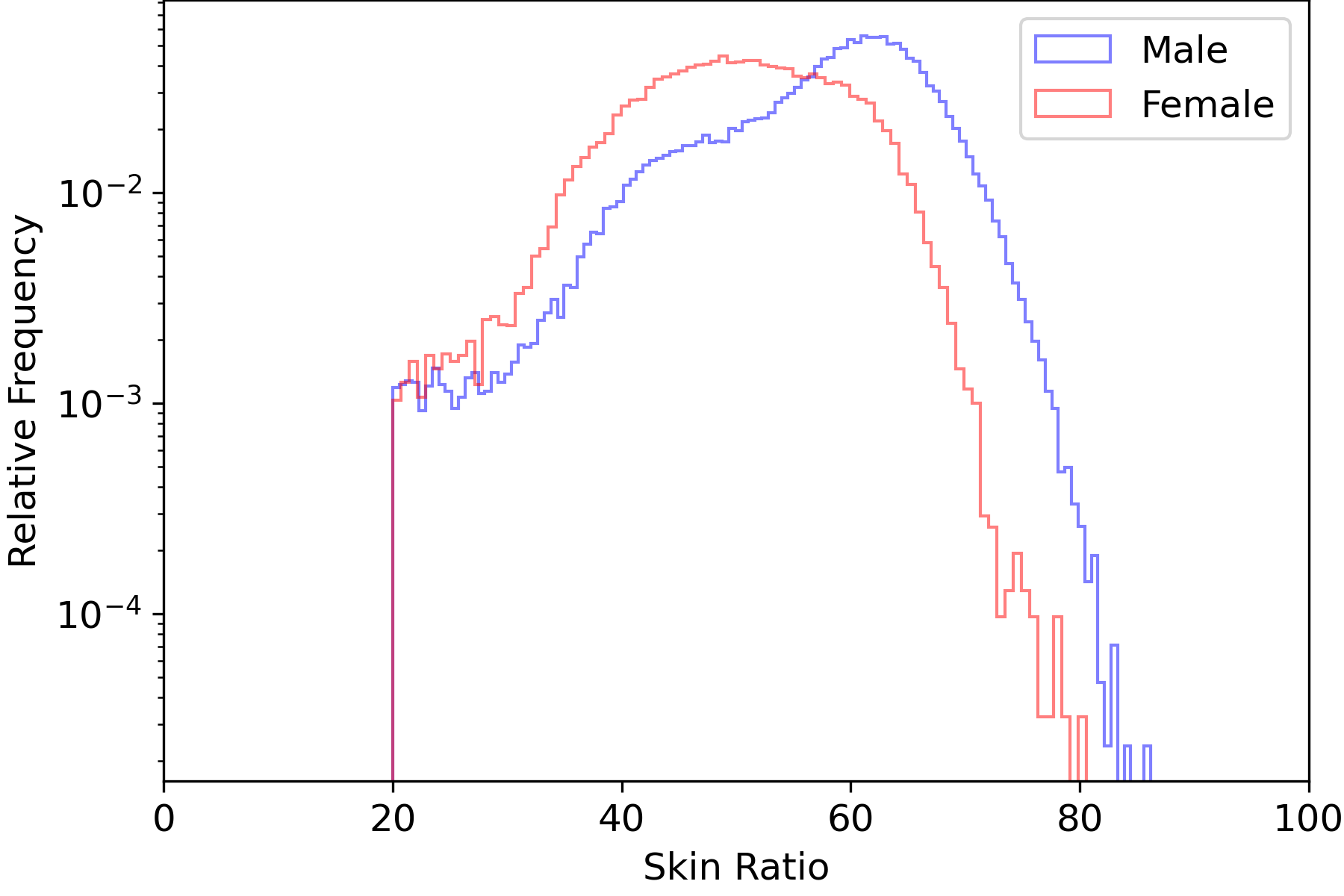}
          \caption{Asian-Celeb}
          \vspace{-0.5em}
      \end{subfigure}
  \end{subfigure}
  \caption{Comparison of female / male distributions of percent of image labelled as face by BiSeNet segmenter.}
  \vspace{-1.0em}
  \label{fig:skin_dist}
\end{figure*}

A different view of this information is the distribution of the fraction of the image that represents the face.
As Asian-Celeb is a web-scraped dataset, and image quality varies (which can affect the face segmentation), we removed images with less than 20\% of the image rated as face, following what was observed in MORPH.
The ``\% face'' distributions compared in Figure~\ref{fig:skin_dist} show that
the proportion of female face images is larger in the range of approximately 25\% to 45\% of the image representing face for MORPH, and 20\% to 55\% for Asian-Celeb, and the proportion of male face images is larger in the range of about 45\% to 70\% of the image representing face for MORPH, and 55\% to 80\% for Asian-Celeb.
Again it is clear that, {\it on average, female face images have less of the image containing information about the face}.


\subsection{Equal Face Info Improves Female FNMR}
\begin{figure*}[t]
  \begin{subfigure}[b]{0.32\linewidth}
      \begin{subfigure}[b]{1\columnwidth}
        \centering
          \includegraphics[width=\columnwidth]{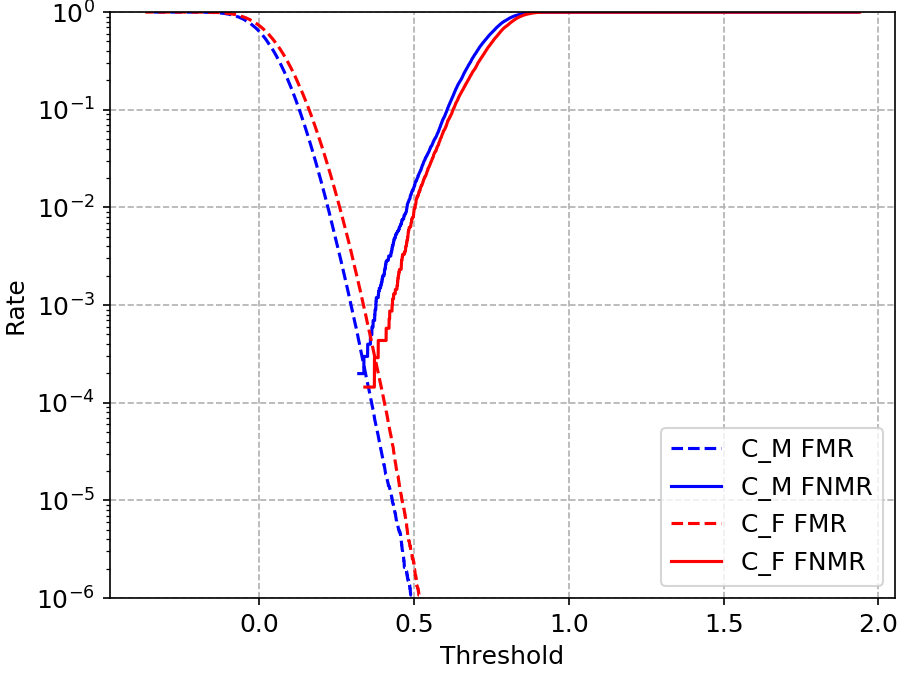}
      \end{subfigure}
      \hfill
      \begin{subfigure}[b]{1\columnwidth}
        \centering
          \includegraphics[width=\columnwidth]{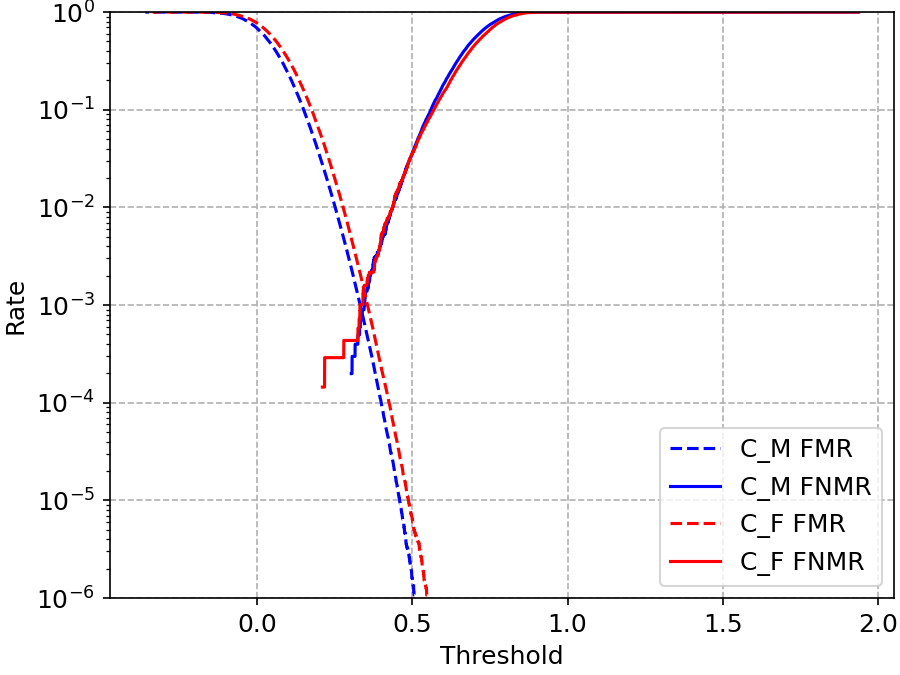}
      \end{subfigure}
      \hfill
      \begin{subfigure}[b]{1\columnwidth}
        \centering
          \includegraphics[width=\columnwidth]{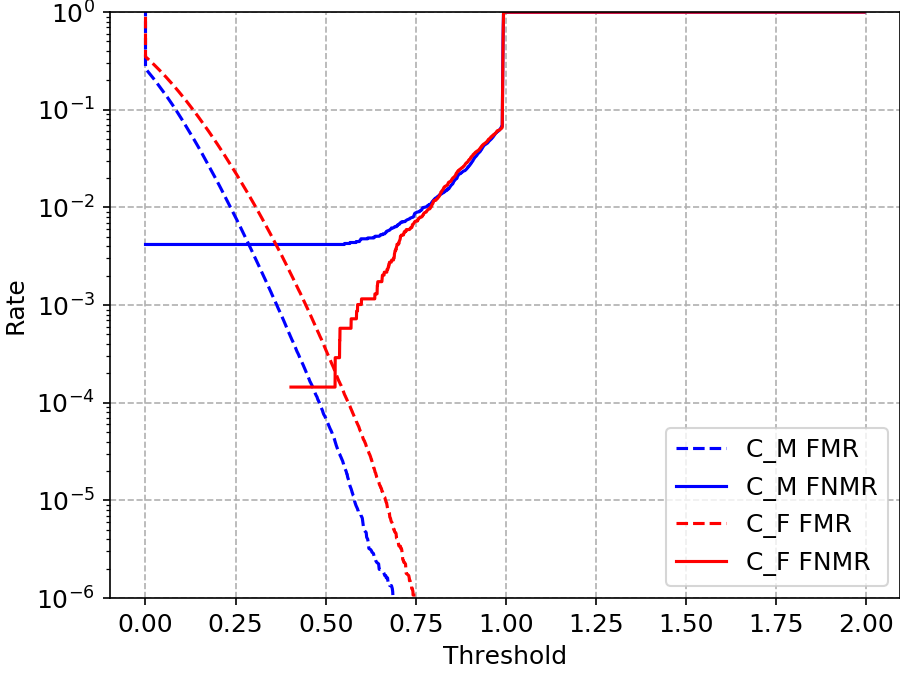}
      \end{subfigure}
      \hfill
      \begin{subfigure}[b]{1\columnwidth}
          \centering
          \begin{subfigure}[b]{0.48\columnwidth}
            \centering
              \includegraphics[width=1\columnwidth]{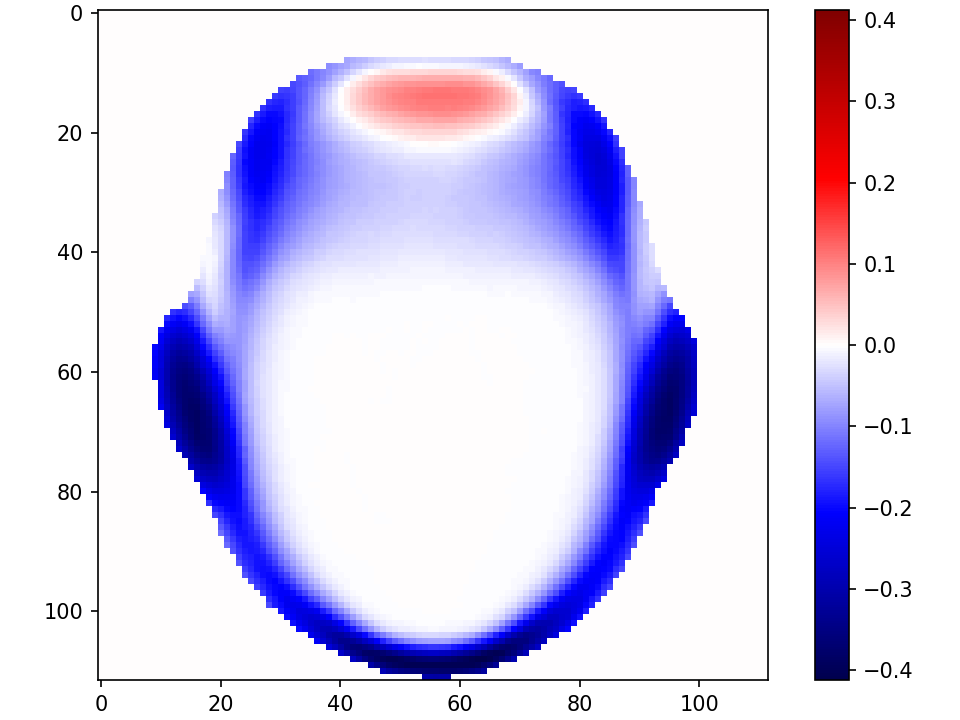}
          \end{subfigure}
          \begin{subfigure}[b]{0.48\columnwidth}
            \centering
              \includegraphics[width=1\columnwidth]{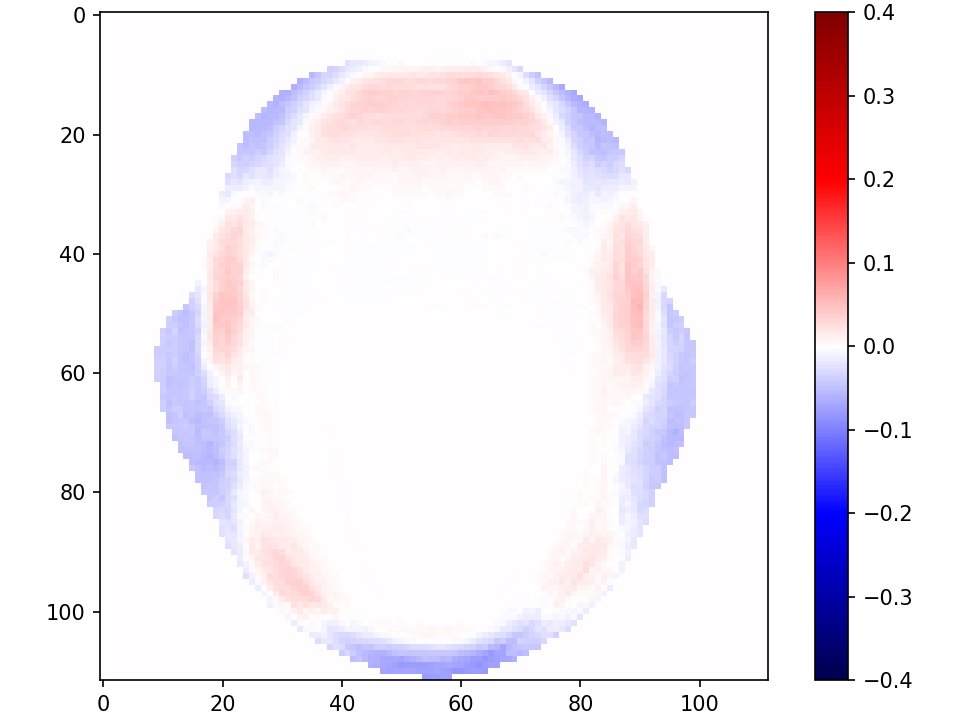}
          \end{subfigure}
      \end{subfigure}
      \caption{MORPH Caucasian}
  \end{subfigure}
  \hfill
  \begin{subfigure}[b]{0.32\linewidth}
      \begin{subfigure}[b]{1\columnwidth}
        \centering
          \includegraphics[width=\columnwidth]{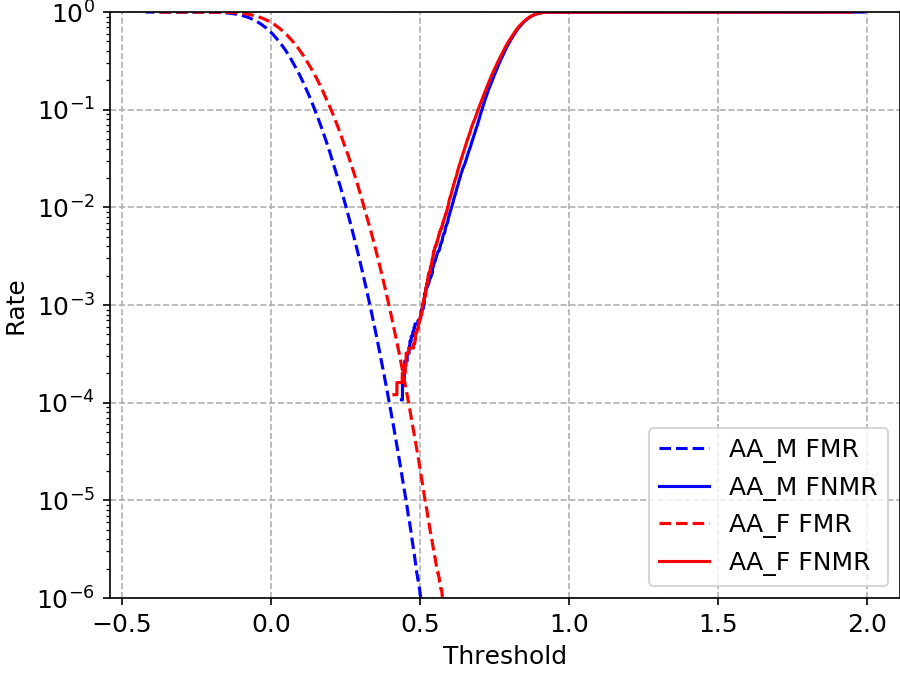}
      \end{subfigure}
      \hfill
      \begin{subfigure}[b]{1\columnwidth}
        \centering
          \includegraphics[width=\columnwidth]{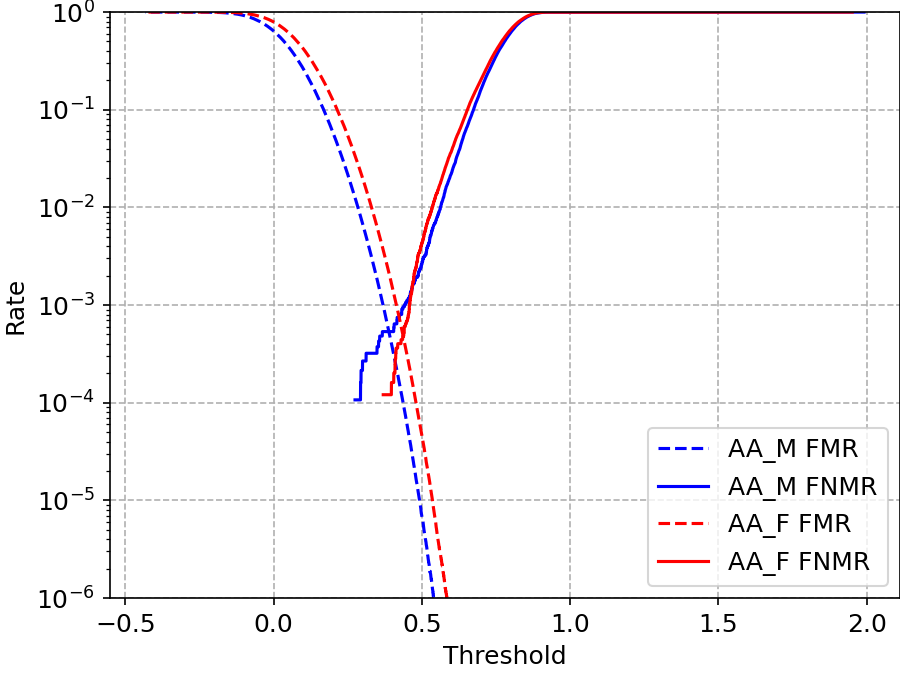}
      \end{subfigure}
      \hfill
      \begin{subfigure}[b]{1\columnwidth}
        \centering
          \includegraphics[width=\columnwidth]{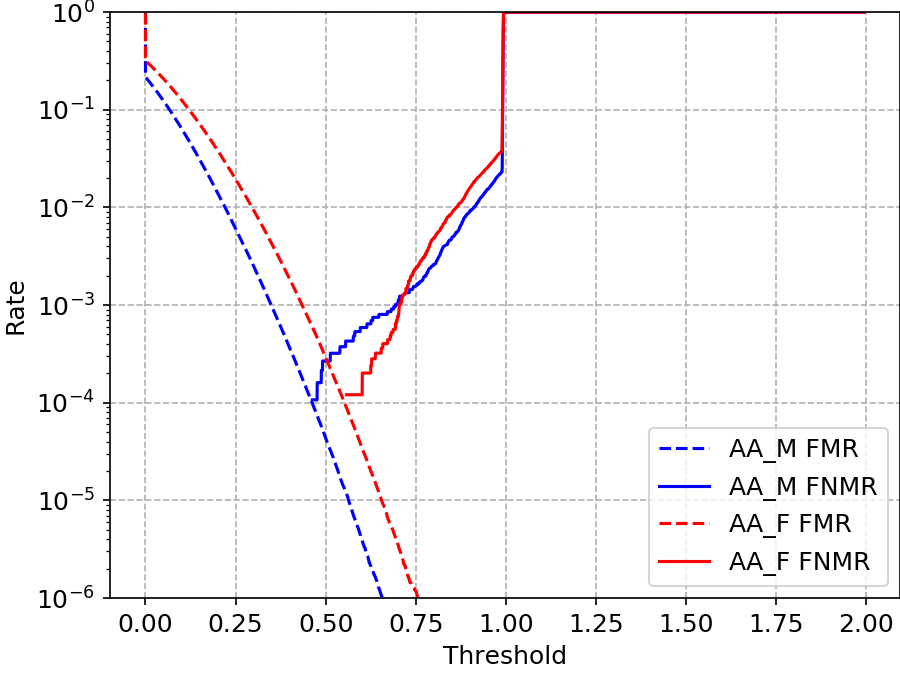}
      \end{subfigure}
      \hfill
      \begin{subfigure}[b]{1\columnwidth}
          \centering
          \begin{subfigure}[b]{0.48\columnwidth}
            \centering
              \includegraphics[width=1\columnwidth]{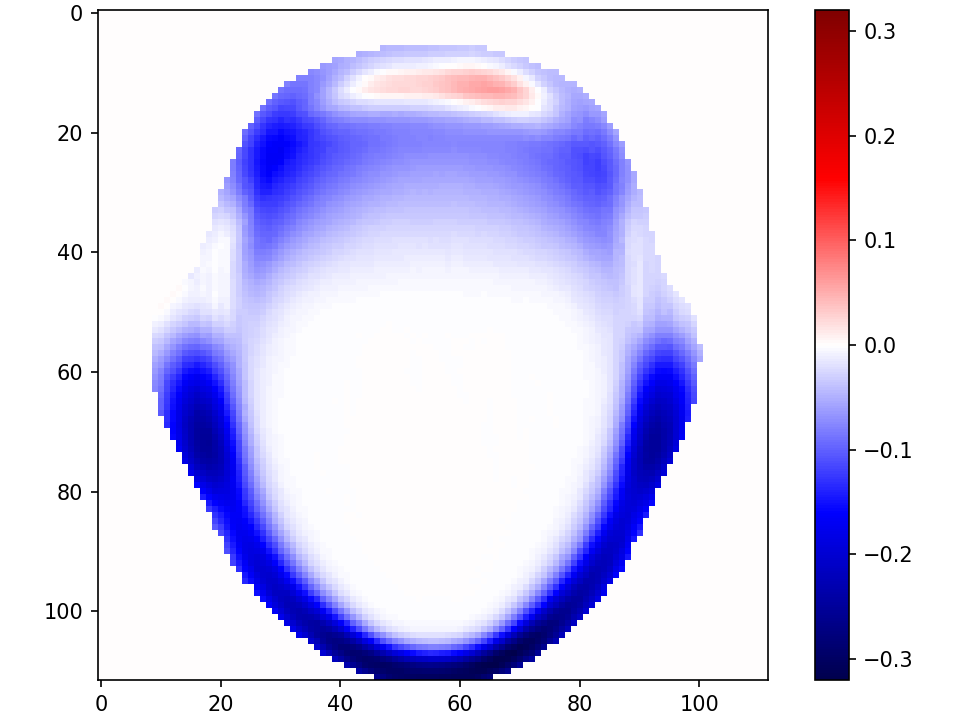}
          \end{subfigure}
          \begin{subfigure}[b]{0.48\columnwidth}
            \centering
              \includegraphics[width=1\columnwidth]{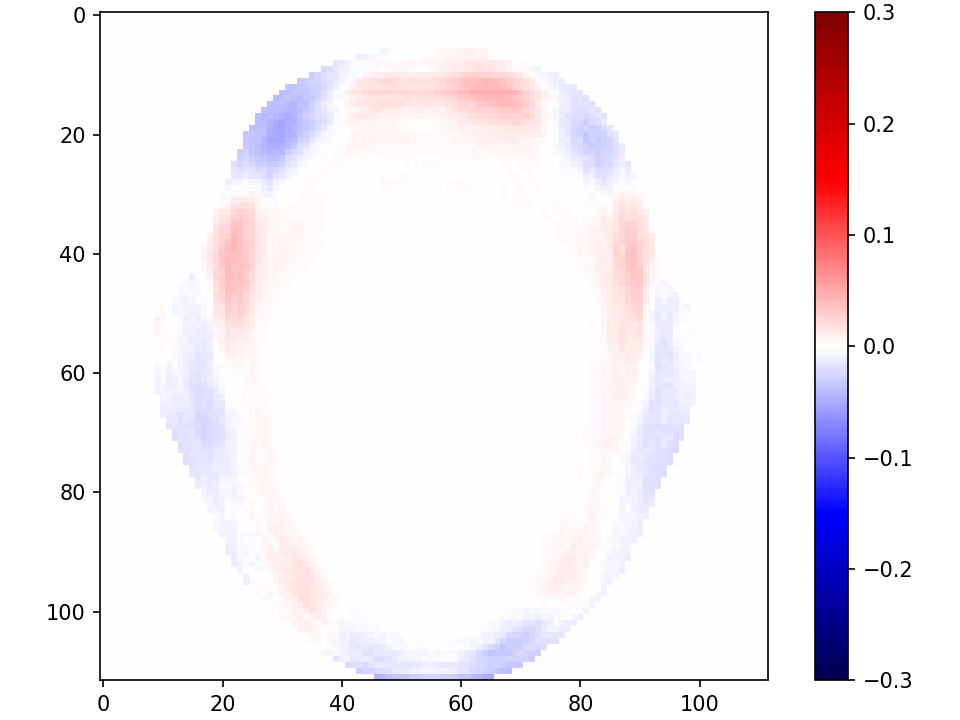}
          \end{subfigure}
      \end{subfigure}
      \caption{MORPH African-American}
  \end{subfigure}
  \hfill
  \begin{subfigure}[b]{0.32\linewidth}
      \begin{subfigure}[b]{1\columnwidth}
        \centering
          \includegraphics[width=\columnwidth]{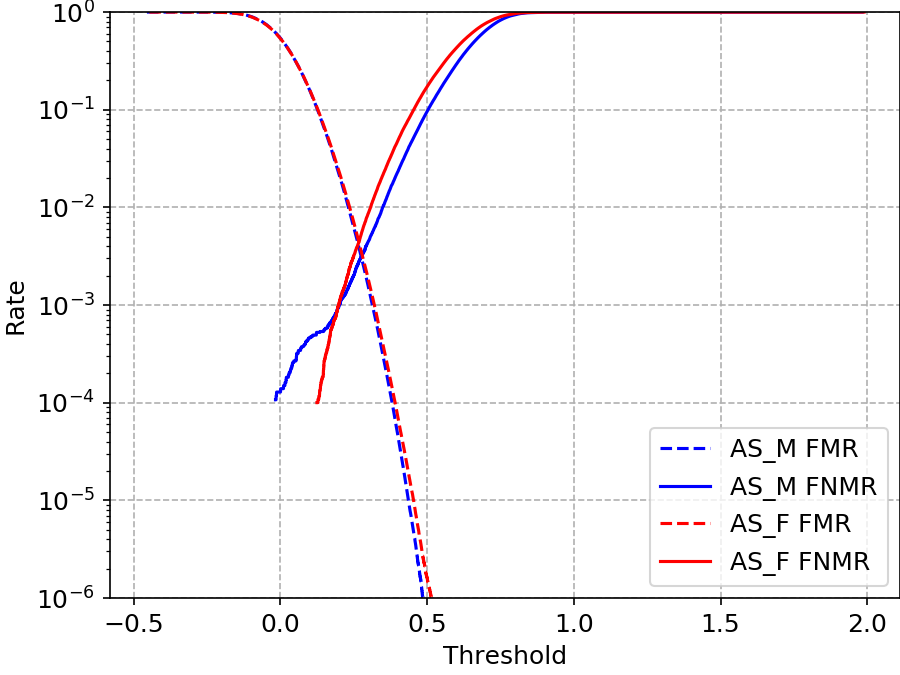}
      \end{subfigure}
      \hfill
      \begin{subfigure}[b]{1\columnwidth}
        \centering
          \includegraphics[width=\columnwidth]{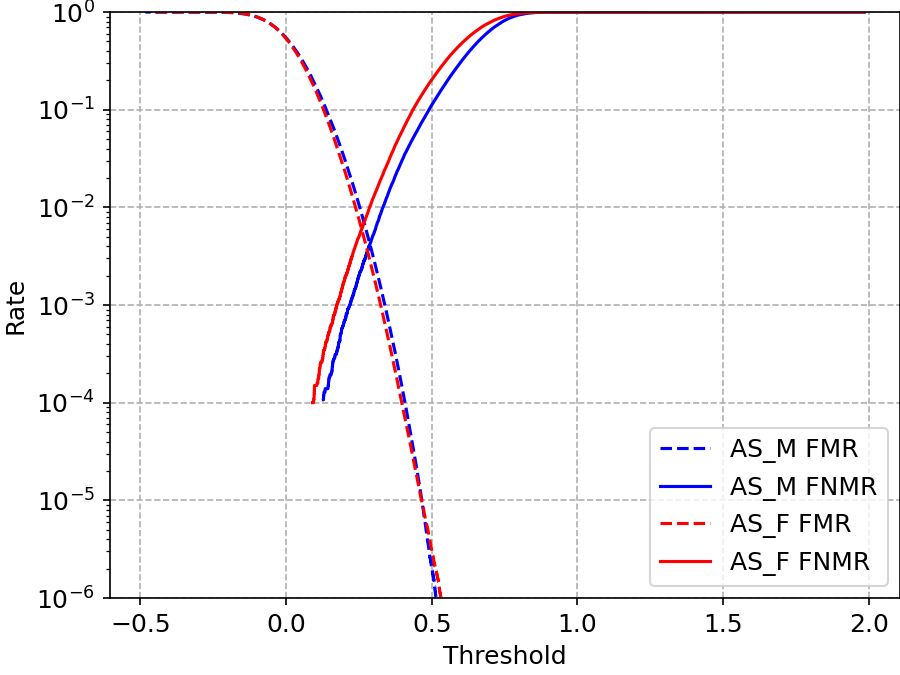}
      \end{subfigure}
      \hfill
      \begin{subfigure}[b]{1\columnwidth}
        \centering
          \includegraphics[width=\columnwidth]{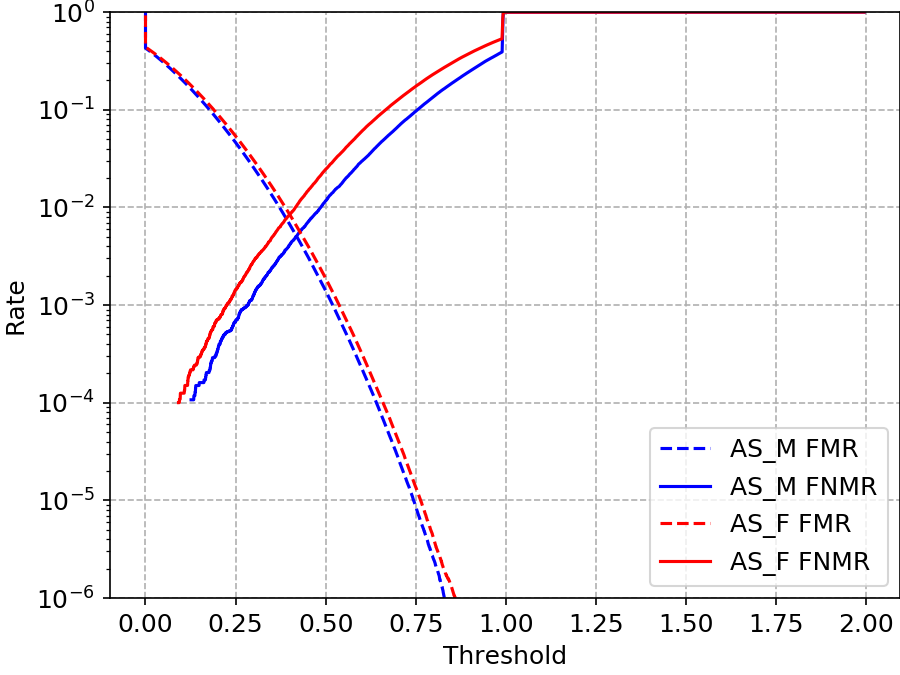}
      \end{subfigure}
      \hfill
      \begin{subfigure}[b]{1\columnwidth}
          \centering
          \begin{subfigure}[b]{0.48\columnwidth}
            \centering
              \includegraphics[width=1\columnwidth]{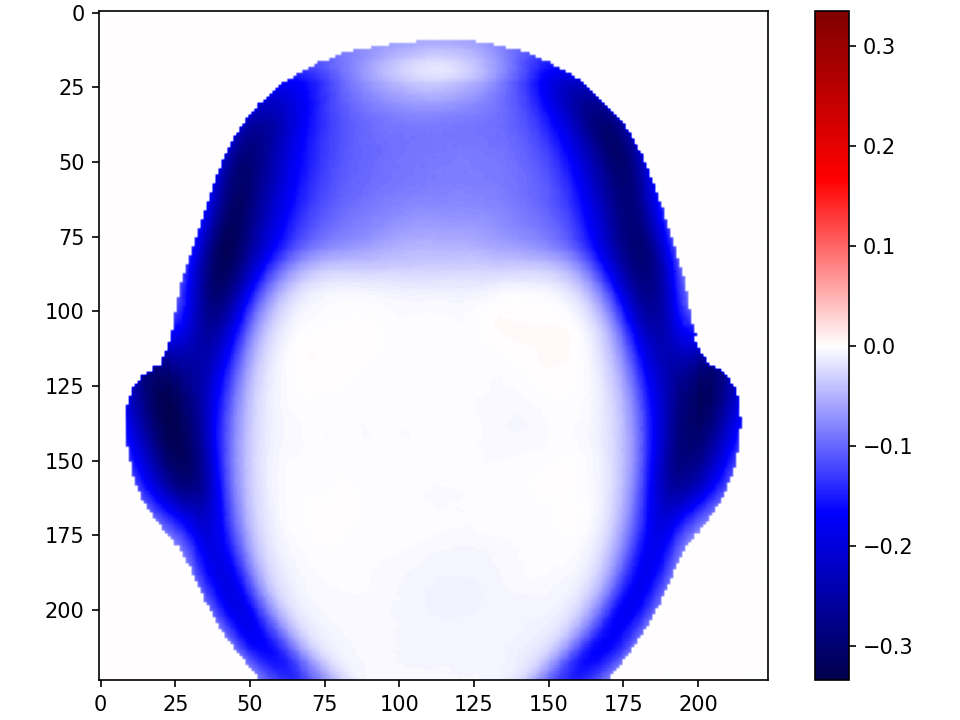}
          \end{subfigure}
          \begin{subfigure}[b]{0.48\columnwidth}
            \centering
              \includegraphics[width=1\columnwidth]{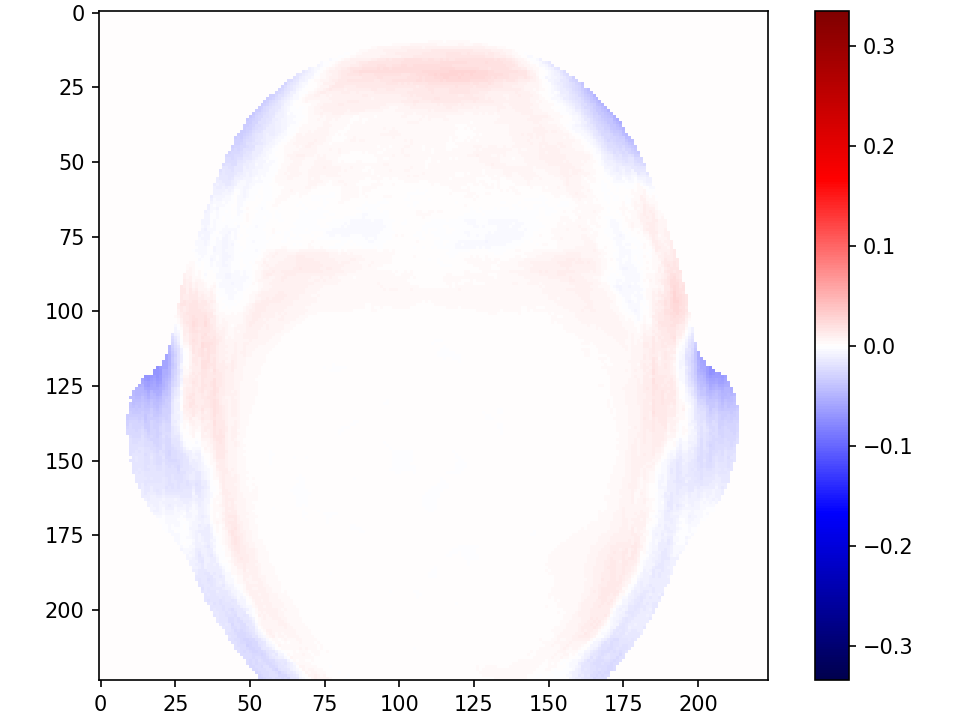}
          \end{subfigure}
      \end{subfigure}
      \caption{Asian-Celeb}
  \end{subfigure}
  \caption{Impostor and genuine distributions for info-equalized female and male image sets. The original image set is first masked at 10\% level from female heatmap, giving the difference heatmap on lower left. A subset of the male images is selected using the intersection-over-union (IoU) to balance with the female images, giving the difference heatmap on lower right. This information-equalized set of images results in the genuine and impostor distributions in the upper rows: ArcFace on top, gender-balanced matcher on middle, and COTS on bottom.}
  \vspace{-1.0em}
  \label{fig:equalized_images}
\end{figure*}

The previous section shows that the observation that face recognition accuracy is lower for females is based on 
images in which the female face images contain less information about the face.
Given this, it is natural to ask what happens if the dataset is controlled to have equal information in female and male face images.

We created a version of the dataset designed to minimize the difference in information between the female and male face images.
First, we masked the area outside of the 10\% level in the female heatmap, in all female and male images.
We choose to mask 10\% as this threshold still capture most of the faces, however, similar results are achieved using a range of threshold values.
Because the female face region is generally a subset of the male,
this establishes the same maximum set of pixels containing face information in all images, and preserves a large majority of the available face information.
However, this still
leaves a bias of greater information content for male images because, as the comparison heatmap shows, any given pixel is generally labeled as face for a higher fraction of the male images than the female images.
Therefore, as a second step, for each female image, we selected the male image that had the maximum intersection-over-union (IoU) of pixels labeled as face in the two images.
The difference heatmap for this nearly info-equalized set of images,
and the resulting FMR and FNMR,
are shown in Figure~\ref{fig:equalized_images}.
These results are for a set of 4,730 Caucasian female images (2,095 subjects) and 4,730 Caucasian male images (2,799 subjects), and 12,383 African-American female images (4,656 subjects) and 12,383 African-American male images (5,414 subjects) for MORPH, and 16,573 Asian female images (4,116 subjects) and 16,573 Asian male images (5,765 subjects) for Asian-Celeb.

The difference heatmaps show that the difference in information between female and male images is effectively minimized.
Remaining differences at any pixel are generally less than 5\% and are roughly balanced between female and male.
The FNMR for the information-equalized images shows a fundamental change from those for the original dataset.
While the FMR distribution for females is still worse than for males, and
for the exception of Asian-Celeb, {\it the FNMR for females is now the same, or even slightly better than, the FNMR for males.}
Giving the fact that Asian-Celeb had the most different FNMR across gender, the skin balancing was still able to push the female and male genuine distributions significantly closer.

Note that this change in the comparison of the female and male FNMR occurs solely from creating an approximately information-equalized test set. The matching algorithm is the same, and was trained on the MS1MV2 dataset which has known female under-representation. 
This is consistent with the experimental observation that gender balance in the training data does not result in gender balance in accuracy on a test set~\cite{Albiero2020_train}.
In effect, this suggests that the common experimental observation that females have a worse genuine distribution is based on a cultural bias, gendered hairstyles, in the test images rather than unbalanced training data.

\begin{figure*}[t]
  \begin{subfigure}[b]{1\linewidth}
      \centering
      \begin{subfigure}[b]{0.327\linewidth}
        \centering
          \includegraphics[width=\linewidth]{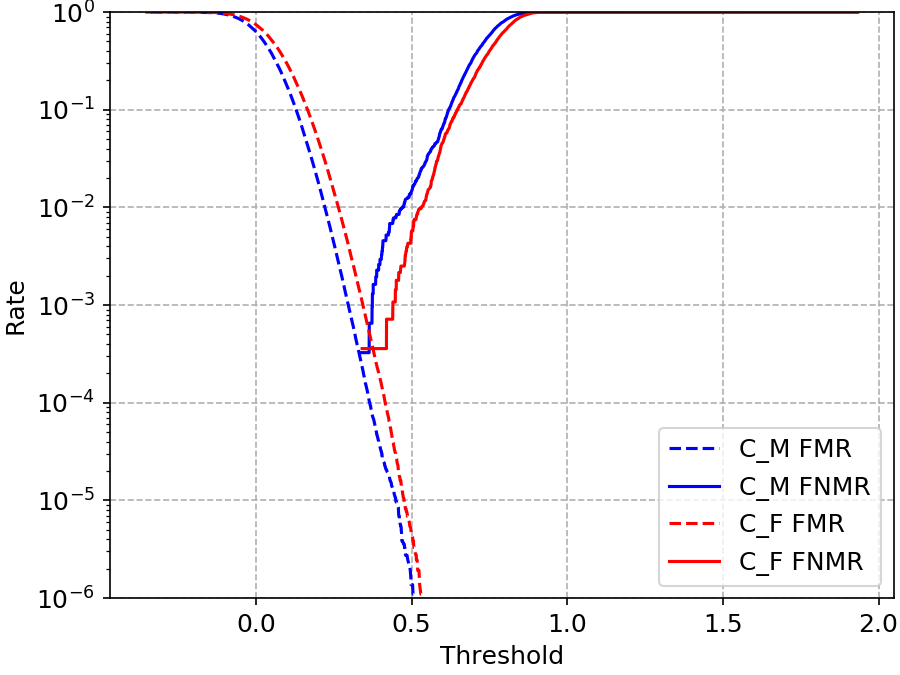}
      \end{subfigure}
      \begin{subfigure}[b]{0.327\linewidth}
        \centering
          \includegraphics[width=\linewidth]{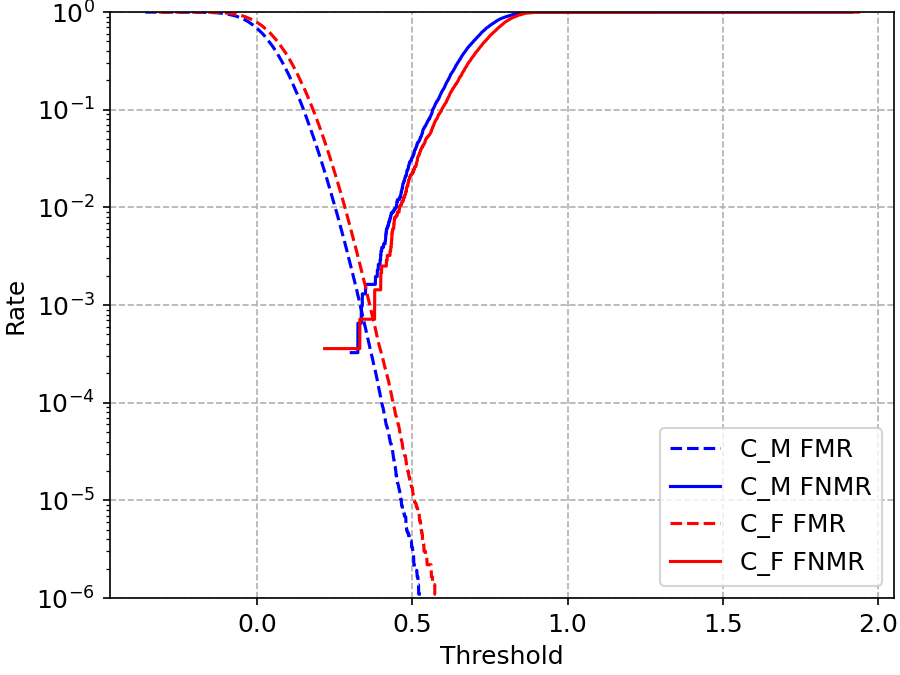}
      \end{subfigure}
      \begin{subfigure}[b]{0.327\linewidth}
        \centering
          \includegraphics[width=\linewidth]{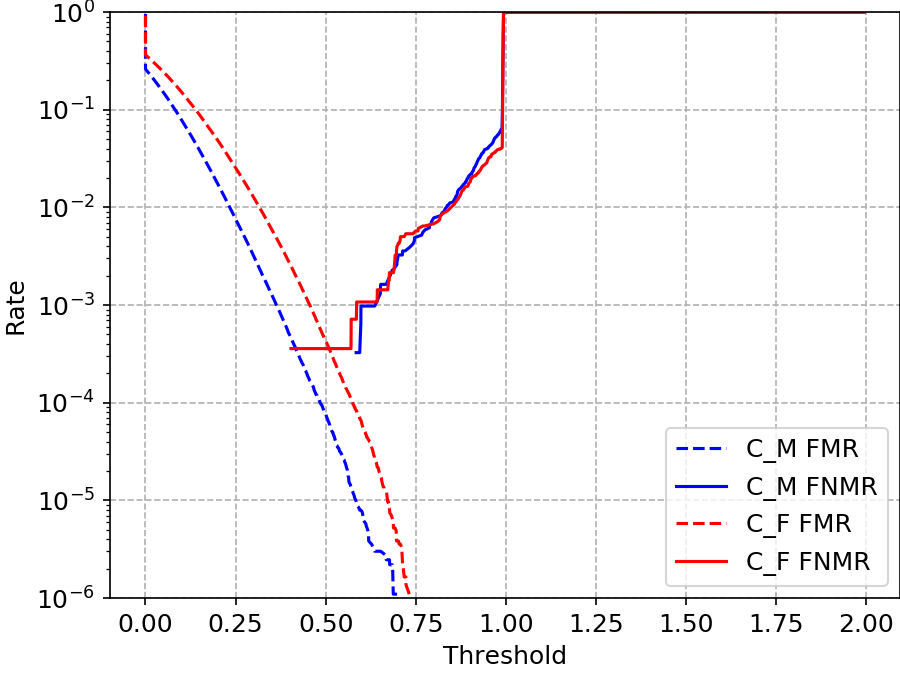}
      \end{subfigure}
  \end{subfigure}
  \begin{subfigure}[b]{1\linewidth}
      \centering
      \begin{subfigure}[b]{0.327\linewidth}
        \centering
          \includegraphics[width=\linewidth]{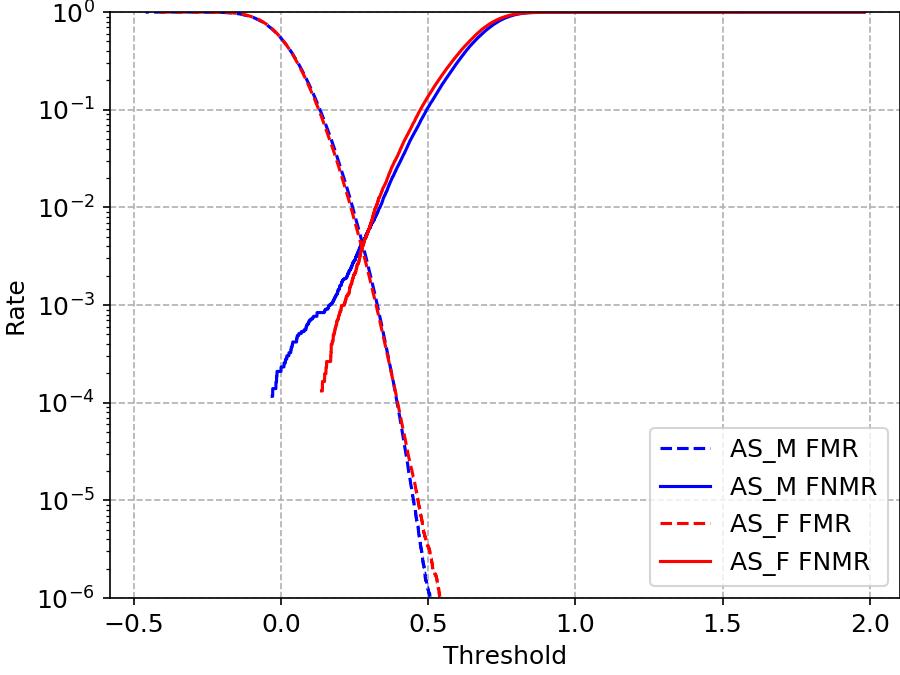}
          \caption{ArcFace}
      \end{subfigure}    
      \begin{subfigure}[b]{0.327\linewidth}
        \centering
          \includegraphics[width=\linewidth]{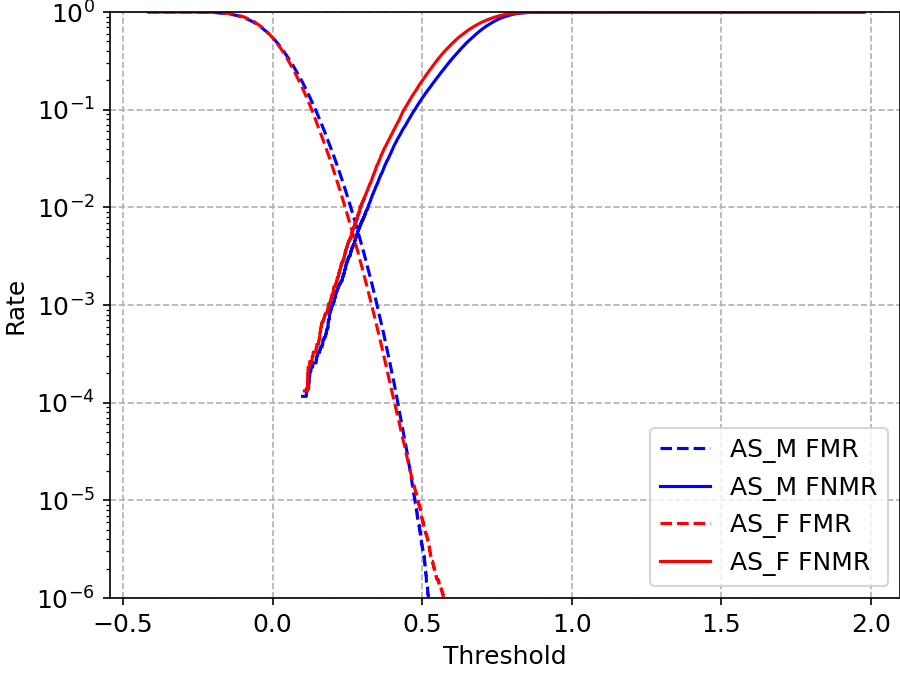}
          \caption{Gender-balanced matcher}
      \end{subfigure}  
      \begin{subfigure}[b]{0.327\linewidth}
        \centering
          \includegraphics[width=\linewidth]{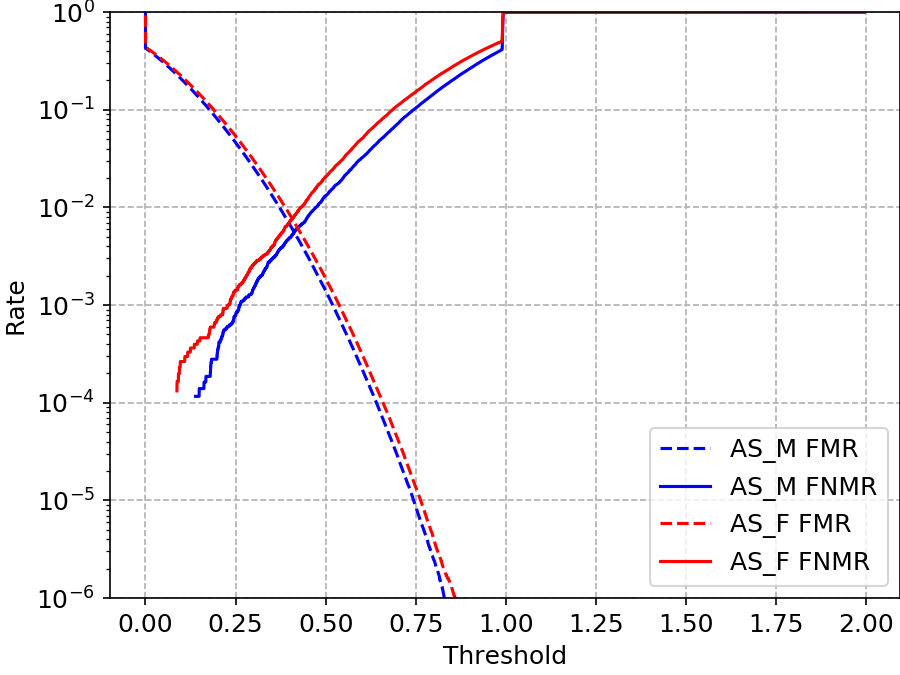}
          \caption{COTS}
      \end{subfigure}    
  \end{subfigure}
  \caption{MORPH Caucasian (top) and Asian-Celeb (bottom) genuine and impostor distribution after skin equalization and makeup balance.}
  \vspace{-1.0em}
  \label{fig:fmr_fnmr_makeup}
\end{figure*}
\begin{figure}[t]
  \begin{subfigure}[b]{1\linewidth}
      \begin{subfigure}[b]{0.32\linewidth}
        \centering
          \includegraphics[width=\linewidth]{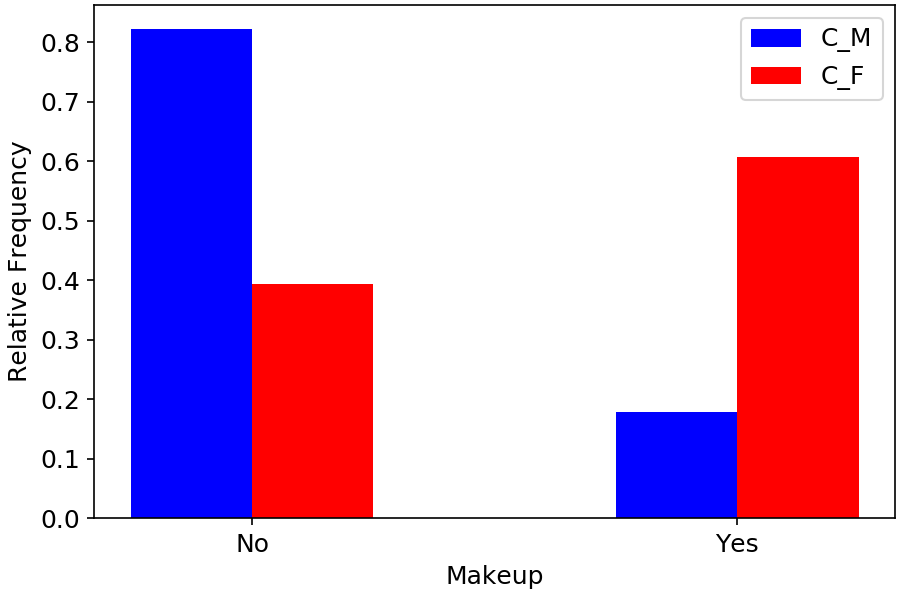}
      \end{subfigure}
      \hfill 
      \begin{subfigure}[b]{0.32\linewidth}
        \centering
          \includegraphics[width=\linewidth]{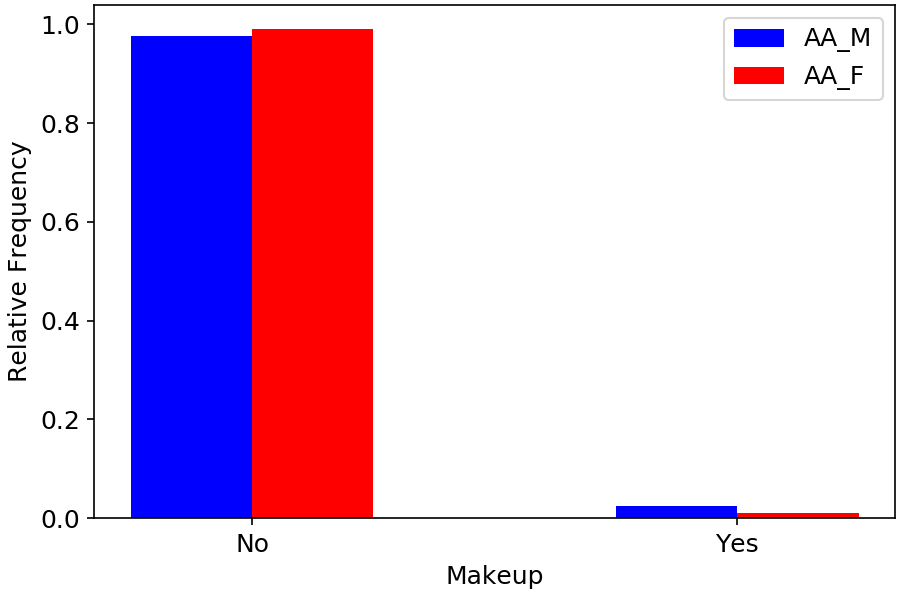}
      \end{subfigure}
      \hfill 
      \begin{subfigure}[b]{0.32\linewidth}
        \centering
          \includegraphics[width=\linewidth]{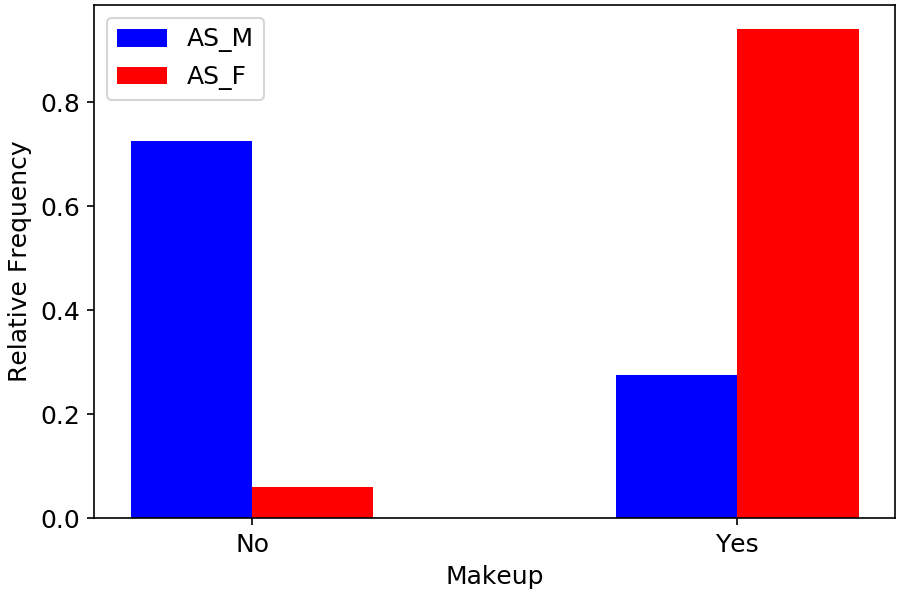}
      \end{subfigure}
  \end{subfigure}
  \caption{Makeup/non-makeup distribution on MORPH Caucasian (left), African-American (middle) and Asian-Celeb (right) datasets using Microsoft Face API.}
  \vspace{-0.5em}
  \label{fig:makeup_dist}
\end{figure}
\subsection{Makeup Effect}

By equalizing the percent of image representing face in male and female images, the FNMR for females is now closer to or sometimes better than for males.
Based on results found by~\cite{Albiero2020_gender}, we investigated what factors could account for the instances where there a gap remains in the FNMR and found makeup to be a factor still affecting FNMR in some instances.

\begin{figure*}[t]
    \begin{subfigure}[b]{1\linewidth}
      \begin{subfigure}[b]{0.327\linewidth}
        \centering
          \includegraphics[width=\linewidth]{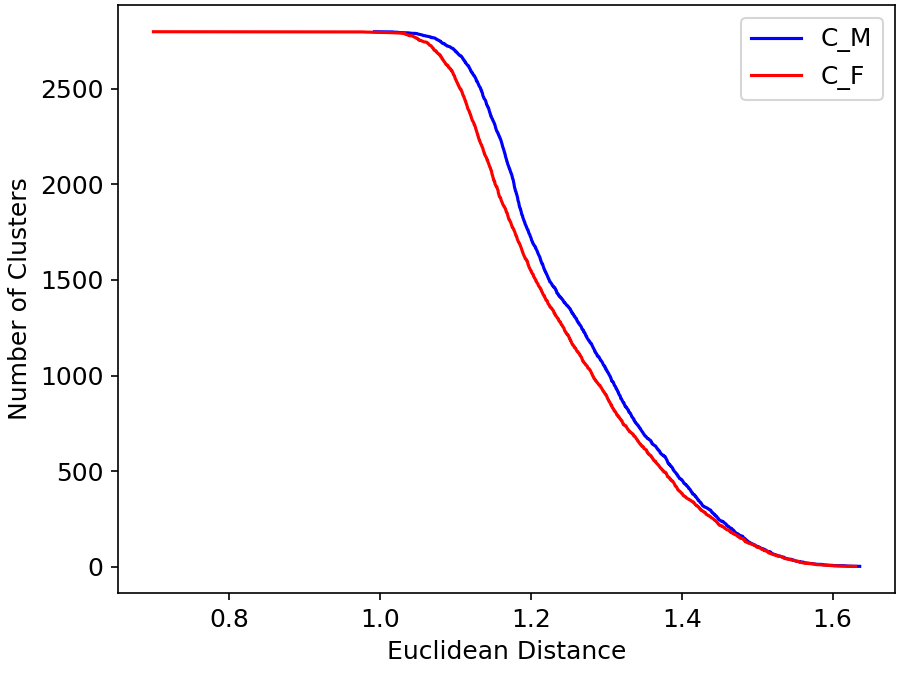}
      \end{subfigure}
      \hfill 
      \begin{subfigure}[b]{0.327\linewidth}
        \centering
          \includegraphics[width=\linewidth]{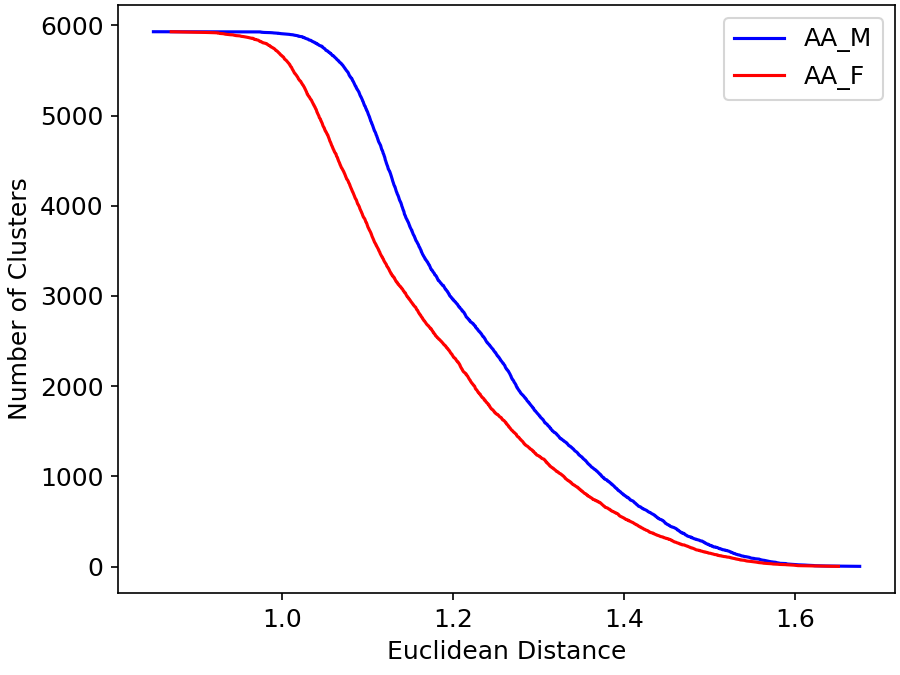}
      \end{subfigure}
      \hfill 
      \begin{subfigure}[b]{0.327\linewidth}
        \centering
          \includegraphics[width=\linewidth]{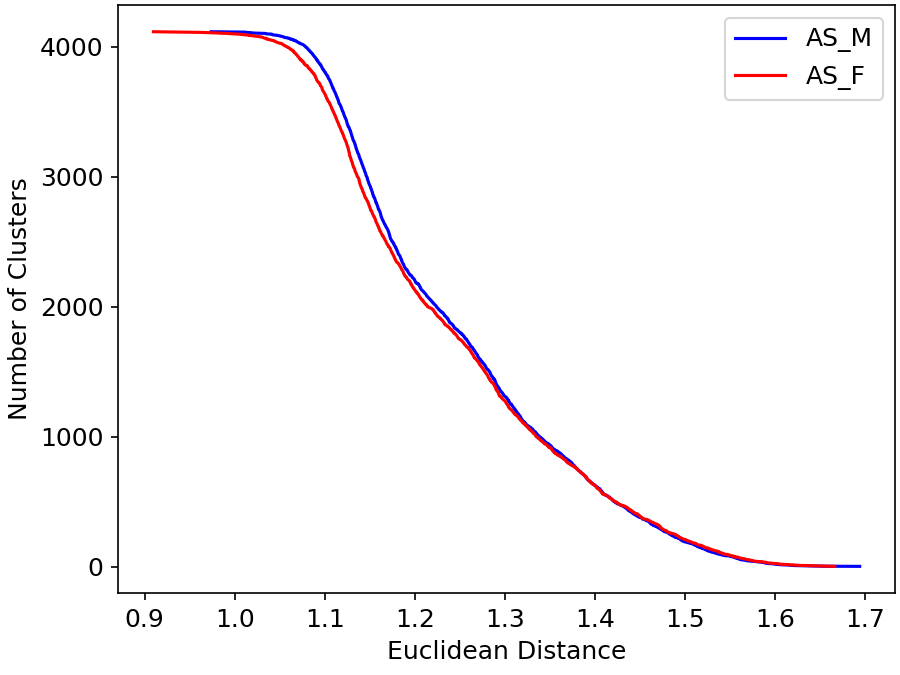}
      \end{subfigure}
  \end{subfigure}
  \begin{subfigure}[b]{1\linewidth}
      \begin{subfigure}[b]{0.327\linewidth}
        \centering
          \includegraphics[width=\linewidth]{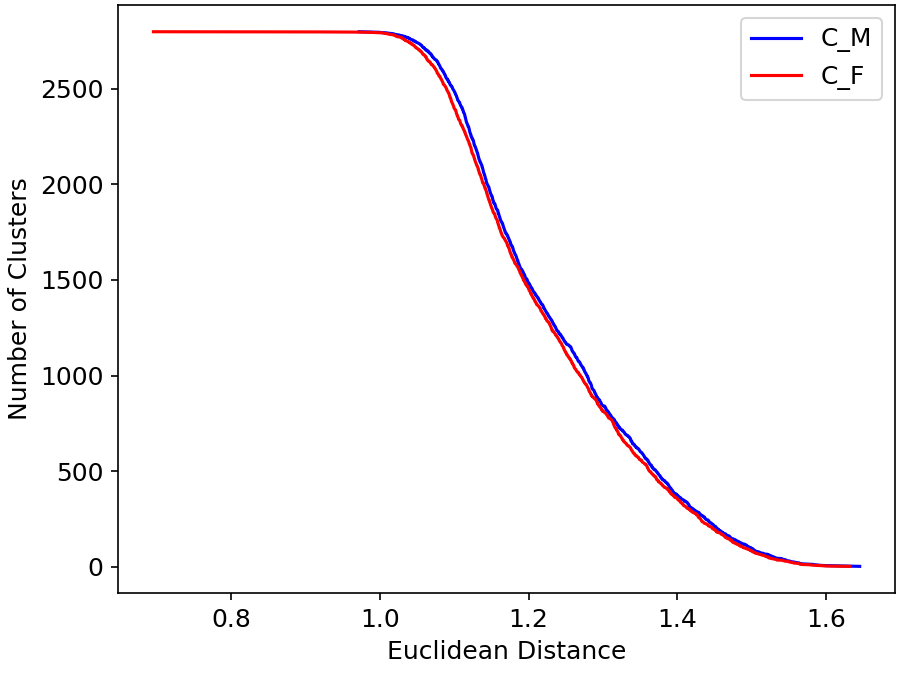}
          \caption{MORPH Caucasian}
      \end{subfigure}
      \hfill 
      \begin{subfigure}[b]{0.327\linewidth}
        \centering
          \includegraphics[width=\linewidth]{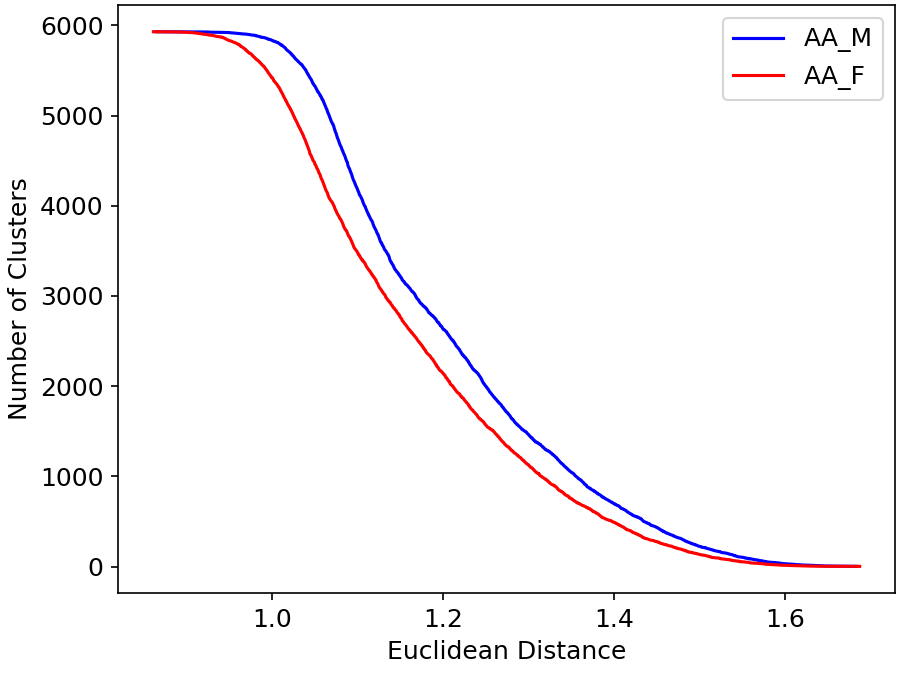}
          \caption{MORPH African-American}
      \end{subfigure}
      \hfill 
      \begin{subfigure}[b]{0.327\linewidth}
        \centering
          \includegraphics[width=\linewidth]{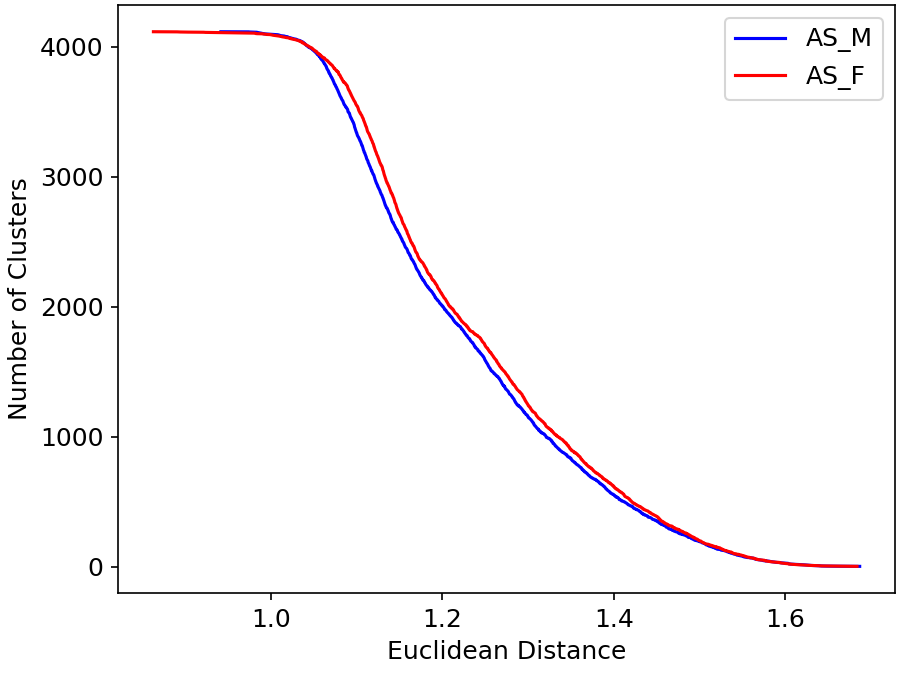}
          \caption{Asian-Celeb}
      \end{subfigure}
  \end{subfigure}
  \caption{Number of clusters as distance threshold increases for ArcFace features (top), and gender-balanced matcher (bottom). Datasets compared have same number of subjects, and one image per subject, thus clusters are formed by different persons.}
  \vspace{-1.0em}
  \label{fig:hier_cluster}
\end{figure*}

Using the Microsoft Face API~\cite{microsoft_api}, we classified the face images as containing or not containing makeup.
Figure~\ref{fig:makeup_dist} shows the distribution of makeup/non-makeup for the three datasets used.
Makeup use is not significant in all instances, but for
 the MORPH Caucasian and Asian-Celeb datasets, females have many more images classified as having makeup than do males.
With this information, we balanced the MORPH Caucasian and Asian-Celeb subsets that were previously skin equalized, so that males and females have the same amount of images classified as makeup/non-makeup.
This leaves us with 2,702 images of 1,458 Caucasian females, and 2,702 images of 1,863 Caucasian males for the MORPH Caucasian dataset, and 7,820 images of 2,881 Asian females, and 7,820 images of 2,777 Asian males.

The FMR and FNMR curves after makeup balancing are shown in Figure~\ref{fig:fmr_fnmr_makeup}.
Comparing with the makeup-unbalanced version, we can observe that FNMRs are even better for females on MORPH Caucasian.
For Asian-Celeb, females are still worse than males, but the difference is now even smaller.
Again, we believe that the fact that this is a web-scrated dataset contributes to it having multiple factors that cannot be easily isolated.

\subsection{Face Clusters and FMR}

The information-equalized dataset shows that females have a FNMR close to or even better than than males, but the FMR for females is still worse.
In other words, for the information-equalized images, on average, matching images of two different females results in a higher similarity value than matching images of two different males.
To explore how and why this can be the case, we analyze impostor subjects by performing agglomerative hierarchical clustering using the complete-linkage clustering algorithm.
The clustering algorithm starts with a list of either ArcFace or gender-balanced matcher 512-d feature vectors, each vector representing a separate identity. 
Initially, the vector for each subject is its own cluster. Then, the clusters are incrementally combined using Euclidean distance until there is only a single cluster.

We computed the clusters separately for females and for males,
using the one image with the highest fraction of pixels labeled as face for each person, with skin pixels outside of a 10\% mask zeroed.
For simplicity of comparison, we randomly selected the same number of males as there are females.
This resulted in 2,798 images each for MORPH Caucasian females and males, 5,929 images each for MORPH African-American females and males, and 4,116 images each for Asian-Celeb dataset.

The resulting clusters can be visualized as a dendogram. 
However, as the number of nodes is very large, no meaningful visual analysis can be performed over the dendogram.
Instead, we analyze the clusters linkage with respect to the Euclidean distance threshold.
This experiment starts by counting every image as its own cluster at zero distance, and then as the distance increases, the dendogram is cut, and the number of clusters is counted just below the cut.

Figure~\ref{fig:hier_cluster} compares the number of clusters formed for females and males as the distance threshold increases.
For all the datasets, males start forming clusters with higher thresholds than females.
Moreover, the difference in the male and female curve is proportional to the difference in FMR between male and female.
For MORPH African-American, where the difference in FMR is larger, females form clusters at much lower distance than males.
For MORPH Caucasian, where the difference in FMR is not as large, the difference in the number of clusters curve is smaller.
Now, for Asian-Celeb with ArcFace, where the FMR is almost the same for males and females, the number of clusters are also similar. With gender-balanced matcher, where females have slightly better FMR, the clustering pattern is slightly flipped as well.

The overall results of this clustering experiment imply that images of two different females are more similar than images of two different males, resulting in them being clustered together at a faster rate than males, which translates into a worse FMR.

\section{Conclusions and Discussion}

The impostor and genuine distributions in Figure~\ref{fig:auth_imp} and the FMR and FNMR curves in Figure~\ref{fig:fmr_fnmr} are representative of the consensus of previous research results that show lower face recognition accuracy for females.
The most extensive study documenting differences in female and male face recognition accuracy, in terms of number of different matchers and datasets considered, is the recent NIST report on demographic effects~\cite{frvt3}.
This report found that females consistently have higher FMR (worse impostor distribution), and generally also have higher FNMR (worse genuine distribution) but that there are exceptions to the higher FNMR generalization.

However, the research results covered in the Related Work section are based on test sets which have not been examined for possible differences between female and male images.
One recommendation for future studies on demographic differences in face recognition accuracy is to include information such as the distributions for the fraction of the image which contains face information, or the face difference heatmap, for the test images, and statistics on the use of makeup, for females and males.
This would encourage an awareness of how test data may come with a built-in ``bias'' toward observing certain accuracy differences.

Differences in size and shape of the female and male face are well known and have been studied in diverse fields 
such as ergonomics~\cite{Zhuang2010}, anthropology~\cite{Holton2014}, and surgery~\cite{Farkas2005}.
(Female and male faces also change differently with age~\cite{Albert2007}.)
For example, measured differences in face size for females and males are reported by Farkas et al.~\cite{Farkas2005}.
For the “North American White Young Adult” persons (Table 1 in~\cite{Farkas2005}) in their study,
they report that the physiognomical face height is 172.5 mm for females versus 187.5 mm for males, 
biocular width (distance between the outer eye corners, or exocanthon) is 86.8 for females versus 89.4 for males, the face width (zygons) is 129.9 for females versus 137.1 for males, and the mandible width is 91.1 for females versus 97.1 for males.
As a matter of facial morphology, the average size and shape of the female face is smaller than that of the male face.
We speculate that facial morphology differences between female and male  give rise to the greater similarity in female impostor pairs that is seen in the clustering analysis in the previous section. 

In initial results, females generally have a worse genuine distribution, and so higher FNMR at the same decision threshold as males.
However, females also generally have different hairstyle than males, and different face shape, and these result in a smaller fraction of the image, on average, containing face information for females.
Also, females generally have greater variation in use of makeup, compared to males.
When the image datasets on which test accuracy is computed are controlled to equalize the fraction of the image that represents face, and for the use of makeup, the female genuine distribution is as good as, or better than, the male genuine distribution.
This is the result seen for the information-equalized dataset in Figure~\ref{fig:equalized_images}.
Thus, the result that females have a higher FNMR than males is found to be caused by a combination of gendered difference in hairstyle, makeup use, and face size and shape.
Because hairstyle conventions, makeup use, and gendered difference in face size and shape may all vary between racial groups, our results and conclusions are consistent with, and may explain, the observation in the NIST report~\cite{frvt3} that the gender difference in FNMR is not universal across different racial groups.

Experimental evidence suggests that the main causes of lower face recognition accuracy for females are gendered social conventions for hairstyle and for makeup, and morphology differences in face size and shape. Some might take the position that the causes are not relevant and that the unequal accuracy must mean that the technology is sexist. Others might take the position that linking cause and effect is essential, and that face recognition technology is not sexist but that the gendered social conventions and biology cause a sexist result. Regardless, the field can and should seek to develop improved face recognition technology that results in more equal accuracy, and a correct understanding of cause and effect may prove useful in this endeavor.

It is reasonable to ask what can be learned from this study of accuracy difference across gender that can help in the study of accuracy difference across race or across age ranges. The causes identified in this study of accuracy difference across gender are gendered social conventions of hairstyle, gendered social conventions of makeup, and biological differences in face size and morphology. Biological differences in face size and morphology across racial groups have been studied before by various authors; e.g., \cite{Farkas2005}. Gendered social conventions of hairstyle and makeup, by definition, can vary greatly between social groups, and so seem likely to arise in a variety of ways. Social conventions of hairstyle and makeup also likely change with a person’s age, and so play a role in understanding how face recognition accuracy varies across age ranges.

One possible direction of future research is to extend the results presented here to additional demographic groups.
Another possible future research direction involves analyzing the effect of beard and mustache on males genuine and impostor distributions.
Moreover, following~\cite{balakrishnan2020towards}, a future work could include the use of synthetic data.
Finally, given the findings of this paper, a potential line of work is to train face matchers to take into consideration the differences in the main factor for lower female accuracy - skin visibility. This trained model could potentially learn to equally match males and females, despite females having less skin visibility.

\ifCLASSOPTIONcaptionsoff
  \newpage
\fi

\bibliographystyle{IEEEtran.bst}
\bibliography{IEEEfull, main}

\begin{IEEEbiography}[{\includegraphics[width=1in,height=1.25in,clip,keepaspectratio]{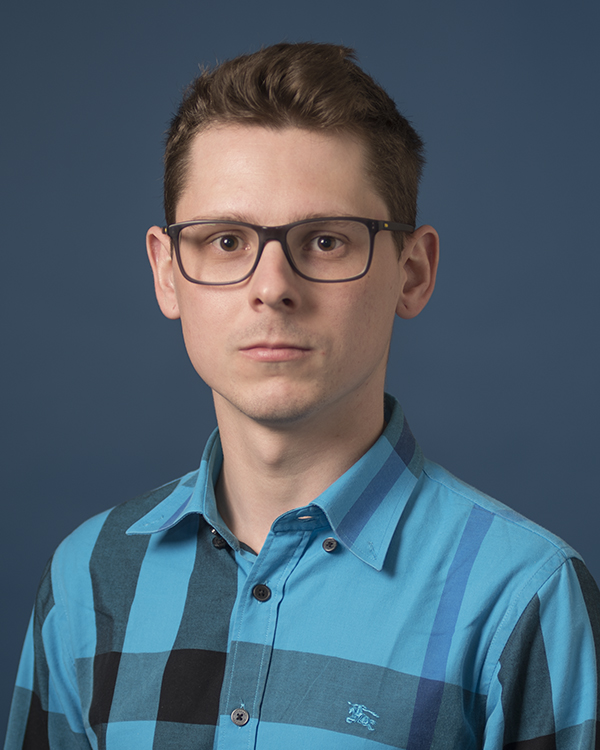}}]{Vítor Albiero (Graduate Student Member, IEEE)}
is a Ph.D. candidate at the University of Notre Dame. He received the B.S. degree in computer science from the University of West Santa Catarina, Brazil, in 2015 and the M.S. degree in Computer Science from the Federal University of Paraná, Brazil, in 2018. He is currently pursuing a Ph.D. degree in Computer Science and Engineering at the University of Notre Dame. His research interests include biometrics, computer vision, and machine learning.
\end{IEEEbiography}

\begin{IEEEbiography}[{\includegraphics[width=1in,height=1.25in,clip,keepaspectratio]{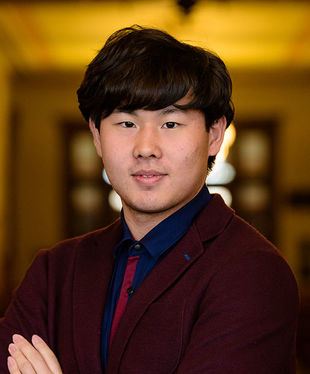}}]{Kai Zhang} is an undergraduate student at the University of Notre Dame. He is currently pursuing a bachelor's degree in Computer Science and Mathematics. His research interests include biometrics, computer vision, and machine learning. 
\end{IEEEbiography}

\begin{IEEEbiography}[{\includegraphics[width=1in,height=1.25in,clip,keepaspectratio]{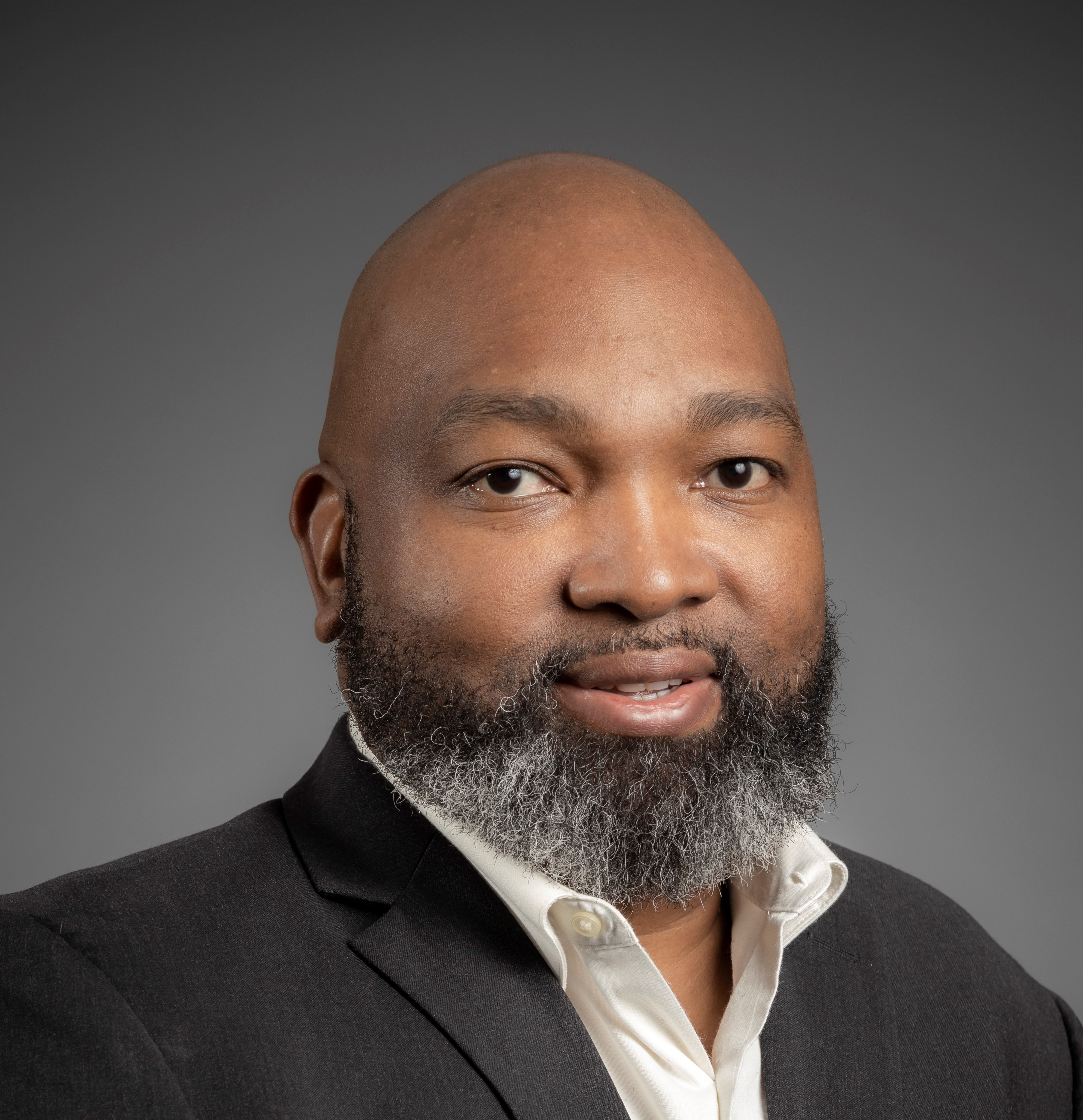}}]{Michael C. King (Member, IEEE)}
joined Florida Institute of Technology’s Harris Institute for Assured Information as a Research Scientist in 2015 and holds a joint appointment as Associate Professor of Computer Engineering and Sciences. Prior to joining academia, Dr. King served for more than 10 years as a scientific research/program management professional in the United States Intelligence Community. While in government, Dr. King created, directed, and managed research portfolios covering a broad range of topics related to biometrics and identity to include: advanced exploitation algorithm development, advanced sensors and acquisition systems, and computational imaging. He crafted and led the Intelligence Advanced Research Projects Activity’s (IARPA) Biometric Exploitation Science and Technology (BEST) Program to transition technology deliverables successfully to several Government organizations.
\end{IEEEbiography}

\begin{IEEEbiography}[{\includegraphics[width=1in,height=1.25in,clip,keepaspectratio]{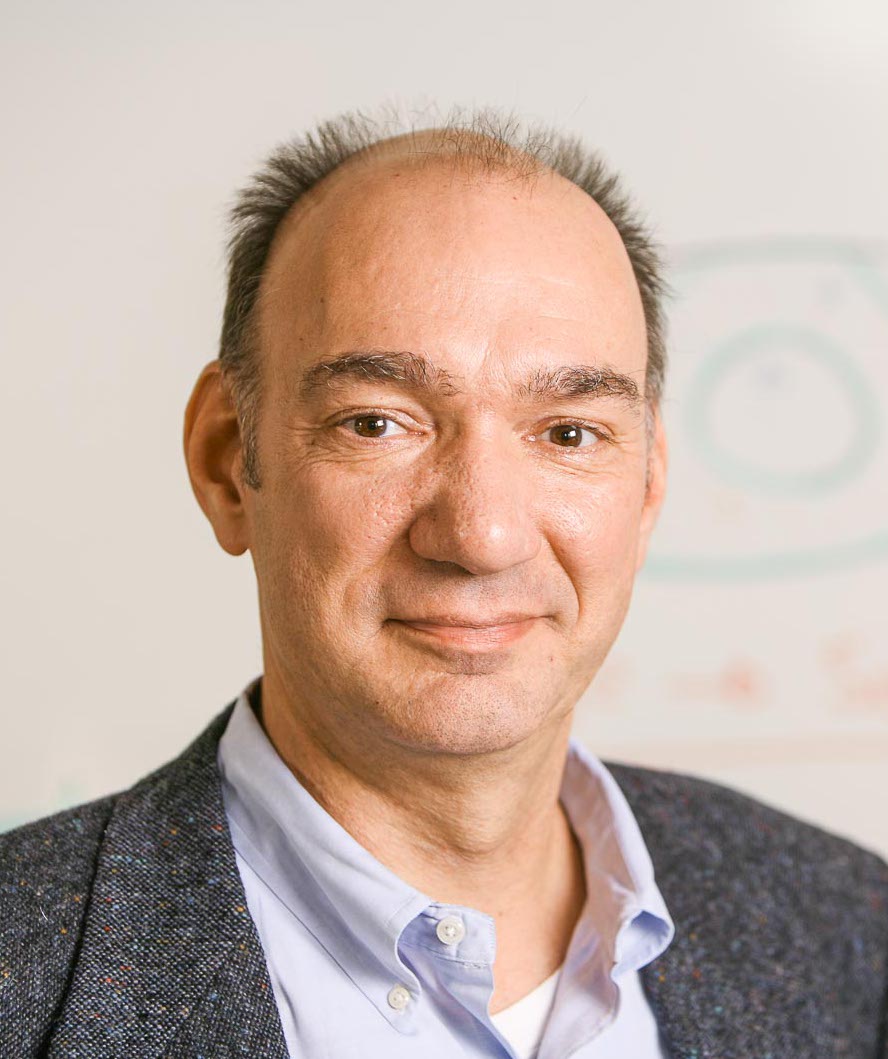}}]{Kevin W. Bowyer (Fellow, IEEE)}
is the Schubmehl-Prein Family Professor of Computer Science and Engineering at the University of Notre Dame, and also serves as Director of International Summer Engineering Programs for the Notre Dame College of Engineering. In 2019, Professor Bowyer was elected as a Fellow of the American Association for the Advancement of Science. Professor Bowyer is also a Fellow of the IEEE and of the IAPR, received a Technical Achievement Award from the IEEE Computer Society, with the citation ``for pioneering contributions to the science and engineering of biometrics''. Professor Bowyer is currently serving as the Editor-in-Chief of the IEEE Transactions on Biometrics, Behavior and Identity Science. He previously served as Editor-in-Chief of the IEEE Transactions on Pattern Analysis and Machine Intelligence.
\end{IEEEbiography}

\end{document}